\documentclass[10pt,twocolumn,letterpaper]{article}

\usepackage{cvpr}              % To produce the CAMERA-READY version
\usepackage{graphicx}
\usepackage{amsmath}
\usepackage{amssymb}
\usepackage{booktabs}
\usepackage{verbatim}

\usepackage[pagebackref,breaklinks,colorlinks]{hyperref}

\usepackage[capitalize]{cleveref}
\crefname{section}{Sec.}{Secs.}
\Crefname{section}{Section}{Sections}
\Crefname{table}{Table}{Tables}
\crefname{table}{Tab.}{Tabs.}

\def\cvprPaperID{*****} % *** Enter the CVPR Paper ID here
\def\confName{CVPR}
\def\confYear{2023}

\newcommand{\mv}[1]{\mathbf{#1}}
\newcommand{\ltil}{\tilde{L}}
\newcommand{\ldet}{\tilde{L}^{\mathrm{(det)}}}
\newcommand{\ocut}{\omega_{\mathrm{cut}}}

\newcommand{\Tcut}{T_{\mathrm{cut}}}
\newcommand{\noff}{N_{\mathrm{OFF}}}
\newcommand{\non}{N_{\mathrm{ON}}}
\newcommand{\nperiod}{N_{\mathrm{period}}}
\newcommand{\coff}{C_{\mathrm{OFF}}}
\newcommand{\con}{C_{\mathrm{ON}}}
\newcommand{\conoff}{C_{\mathrm{ON/OFF}}}

\begin{document}
\pdfsuppresswarningpagegroup=1
\title{Frequency Cam: Imaging Periodic Signals in Real-Time}

\author{Bernd Pfrommer\\
% Unaffiliated\\
{\tt\small bernd.pfrommer@pfrommer.us}
}
\maketitle

%%%%%%%%% ABSTRACT
\begin{abstract}
  Due to their high temporal resolution and large dynamic range event
  cameras are uniquely suited 
  for the analysis of time-periodic signals in an image. In this work
  we present an efficient and fully asynchronous event camera
  algorithm for detecting the fundamental frequency at which image
  pixels flicker. The algorithm employs a second-order digital
  infinite impulse response (IIR) filter to perform an approximate
  per-pixel brightness reconstruction and is more robust to
  high-frequency noise than
  the baseline method we compare to. We further demonstrate that using
  the falling edge of the signal leads to more accurate period
  estimates than the rising edge, and that for certain signals interpolating the
  zero-level crossings can further increase accuracy. Our experiments find that the
  outstanding capabilities of the camera in detecting frequencies up
  to 64kHz for a single pixel do not carry over to full sensor imaging as readout
  bandwidth limitations become a serious obstacle. This suggests that
  a hardware implementation closer to the sensor will allow for
  greatly improved frequency imaging. We discuss the important design
  parameters for full-sensor
  frequency imaging and present Frequency Cam, an open-source implementation as a
  ROS node that can run on a single core of a laptop CPU at more than 50
  million events/sec. It produces results that are qualitatively very similar
  to those obtained from the closed source vibration analysis module in
  Prophesee's Metavision Toolkit. The code for Frequency Cam and a
  demonstration video can be found at \url{https://github.com/berndpfrommer/frequency_cam}.
\end{abstract}

%%%%%%%%% BODY TEXT
%-------------------------------------------------------------------------
\section{Introduction}
\label{sec:introduction}
Unlike traditional frame-based imaging devices, event based cameras~\cite{lichtsteiner_posch_delbrueck}
imitate a biological retina by immediately providing a signal whenever the
illumination of a pixel changes by more than a certain
threshold. Avoiding frame accumulation results in sub-millisecond latencies\cite{hu_liu},
avoids transmission of repetitive information, and lowers power
consumption. Moreover since each pixel adapts to its individual
illumination and emits events based on the logarithm of the photo
current, event cameras by design feature a high dynamic range.

These exciting properties have attracted much interest from the
research community, an overview of which can be found in
Ref.~\cite{gallego_delbrueck_orchard}. A great deal of effort has
focussed on accomplishing high-level tasks in robotics and
computer vision among which are visual odometry~\cite{rebecq_evo}, event-based
classification and object dectection~\cite{schaefer_aegnn}, object
tracking~\cite{zhang_dong} and control~\cite{delbruck_lichtsteiner_control}.

\begin{figure}[t]
  \centering
  \includegraphics[width=0.6\linewidth]{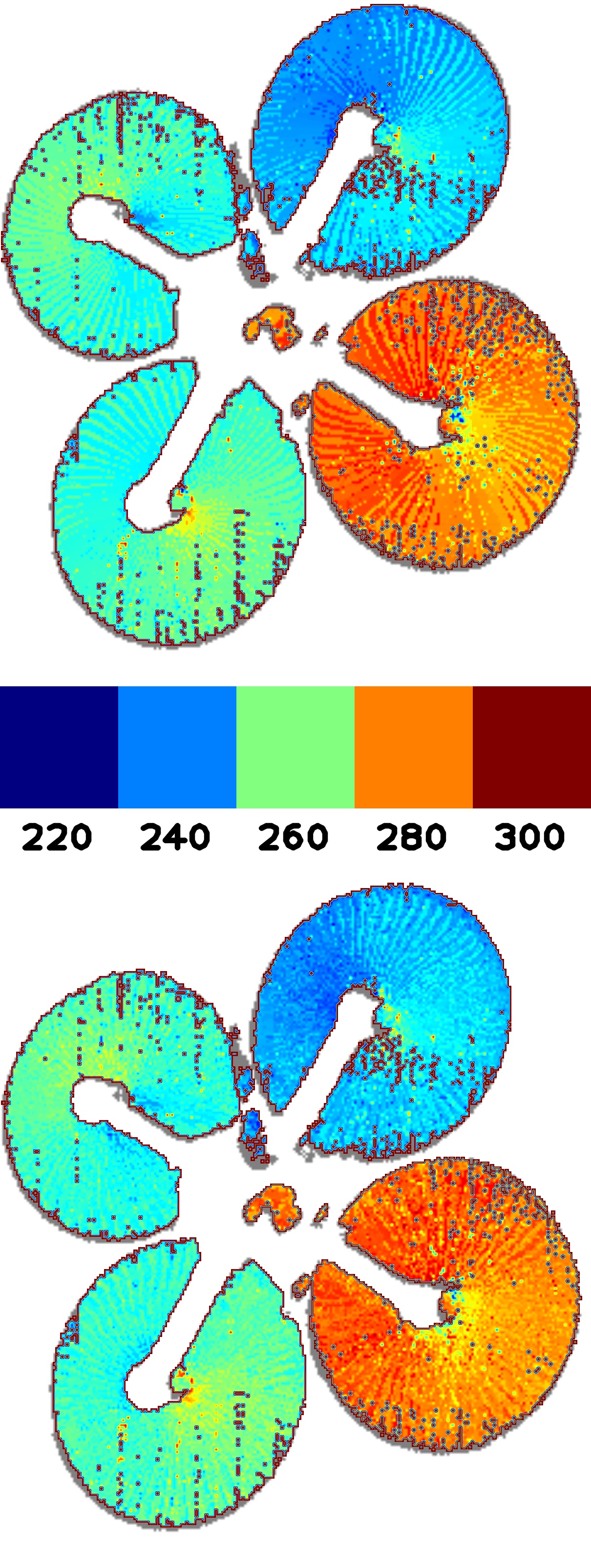}
  \caption{Frequency image of a quad rotor in flight, with
    color coded frequencies (in Hz). The top image was obtained with
    a filter of $\Tcut=5$ whereas the bottom is generated using
    Prophesee's proprietary vibration analysis module. The
    image has been centered and cropped to a size of 242 x 281 for
    ease of viewing. Grey color indicates pixels for which no
    frequency could be determined, but that have had events during the
    readout period (10ms). Both images are qualitatively very similar.}
   \label{fig:quad_freq}
\end{figure}

The present work addresses a much more basic task that has received
surprisingly little attention: how to detect the frequency of
periodically varying signals on a per-pixel level.
Often periodic flickering is considered a nuisance signal to be
filtered out~\cite{wang_yuan}, however there is commercial interest
for frequency detection in the context of vibration analysis. In fact Prophesee's
Metavision Source Development Kit (SDK) contains a vibration analysis
module to which we will compare (Fig.~\ref{fig:quad_freq}).

At this point it is necessary to discuss more precisely what
frequency detection means. For brevity, in the context of this paper
frequency generally refers to the  lowest frequency present in a
periodic signal, also known as the 
fundamental frequency. Proper detection of the fundamental frequency
requires reconstructing the brightness of each individual
pixel (c.f.~Sec.\ref{sec:digital_filter}), storing it in a
suitably large buffer, computing its 
Fourier transform, and finding the lowest-frequency peak in the
spectrum. For off-the-shelf hardware this approach is prohibitively
expensive in terms of memory and compute resources and yet will still
fail in some situations, for example when a high frequency signal is modulated with low
frequency. Such periodogram based methods are also a poor match for the
neuromorphic paradigm of the camera due to the buffering involved.

Some compromises to the objective must be made so it can be tackled by
an asynchronous and light-weight algorithm.
It turns out that many simple signals such as square,
triangular, or sine waves are repetitions of one monotonically rising and one falling
section. In this case one can detect the onset of the falling
section when an ON (illumination decrease) event is followed by an OFF
(illumination increase) event\cite{censi_strubel}. The time period between detecting
successive transitions can then serve as a proxy for the fundamental
frequency. This algorithm has also recently been used for
detecting strobe light signals with a known frequency
~\cite{stoffregen} and will serve as a baseline here.

The present work improves on the robustness of the baseline by
performing an approximate reconstruction of the brightness from the
events by means of a digital filter, followed by the detection of
zero-level crossings. While detecting zero-level crossings is still by no
means the same as finding the fundamental frequency, the approach is
substantially more robust to noise events than the baseline. It no
longer requires the signal to consist of monotonically rising and falling
sections but only mandates that the signal cross the zero level twice
per period. Our approach
further retains the fully asynchronous and light-weight nature of the
baseline, and for some signals allows for more accurate frequency
measurements by interpolation of the zero-level crossing times (Fig.~\ref{fig:triangle_wave}).

As we turn our attention from frequency detection for a single-pixel to full-sensor
frequency imaging we discover:
\begin{itemize}
  \item the severe limitations imposed on the camera's performance by
    finite readout bandwidth,
  \item how strongly lens flare affects frequency detection in scenes
      with high dynamic range,
  \item what is important for an algorithm to produce visually
    pleasing real-time frequency images and,
   \item why the baseline algorithm works surprisingly well in a
     full-sensor setting.
\end{itemize}

The results of our experiments suggest that frequency imaging can
greatly benefit from a hardware implementation close to the sensor.

%-------------------------------------------------------------------------
\section{Related Work}
\label{sec:related}
Before discussing the few
works~\cite{censi_strubel}\cite{hoseini_linares}\cite{hoseini_orchard}
that use event based cameras for frequency detection or estimation we
will set our paper in the context of frequency estimation techniques
used in other areas of signal processing.

% Batch/FFT approaches
As already indicated in Sec.~\ref{sec:introduction} we rule out
methods based on periodograms that involve 
Fourier transforms because they require buffering and do not meet our
goals with respect to latency and compute effort. For this reason we
will limit our attention to so-called parametric approaches which
assume the signal to contain at most a few frequencies.

% Radar algorithms and Hilbert filter
A vast number of parametric algorithms have been developed in the
context of Doppler radar techology and for other sensor arrays, one of
the most successful ones being the MUSIC~\cite{music_schmidt}
algorithm. For single-frequency signals, frequency and phase estimation
can be viewed as a linear regression of the phase
data~\cite{kay}\cite{tretter}, allowing for fast and accurate
algorithms to extract phase and frequency of a base signal under
moderate noise via phase estimation and
unwrapping\cite{slocumb_kitchen}. For all these methods the signal is
assumed to be available in its complex ``analytic`` representation,
i.e. as phase and amplitude. Unfortunately event based cameras do not
present signals as phase and amplitude but just emit events that can be used for
brightness reconstruction to yield at best the (real valued)
amplitude. The conversion to an analytic signal requires either an
undesirable batch approach such as non-linear least squares
(NLS)~\cite{besson_nls} or the design of an infinite impulse response (IIR) Hilbert
transformer, also known as a quadrature filter in the digital signal
processing literature\cite{hilbert_transformers_ansari}. We
pursued but abandoned this direction. Despite much effort we
were unable to design an IIR quadrature filter sufficiently
robust to signal noise as observed in our experiments. We suspect
this also to be an issue with online NLS-style
estimation schemes~\cite{kusljevic_recursive_freq_est} but we did not
explore those.

% Hilbert-Huang
Based on our negative experience with the Hilbert transform we
steered clear of ``instantaneous'' frequency estimation via the
Hilbert-Huang tranform~\cite{hilbert_huang} which is highly popular for
frequency estimation in fields ranging from biomedicine and neuro
sciences to ocean engineering and speech recognition. Besides our
reservations about the Hilbert transform, the empirical mode
decomposition preceeding the Hilbert transform appears prohibitively
compute heavy for our application.

% Power grid
Frequency estimation is also a central topic for electric power grid
monitoring and management. Due to the low frequency (50-60Hz) and because
typically only a small number of signals need to be monitored, batch
methods like periodgrams can be applied. Some online
methods~\cite{li_teng_power_grid} look promising but are tested only
on relatively low-noise data and for the small frequency band within
which the signal is expected to fluctuate. We do however use an idea
that is well known in the power grid
literature~\cite{mendonca_power_grid}: the accuracy of zero-level
crossings can be improved by interpolation. In contrast to
Ref.~\cite{mendonca_power_grid} we avoid running a
linear regression and simply use the nearest neighbor sample points
before and after the crossing.

% Audio and Kalman filter
Pitch detection for audio analysis and speech recognition is often
performed with FFT based algorithms, although there are some
online approaches as for example Kalman-filter based
methods\cite{shi_nielsen_kalman}. We briefly attempted Kalman
filtering for our application but found that updating the filter state
and pertaining matrices creates more memory traffic than a typical
laptop CPU can handle. Moreover the Kalman filters employed are
nonlinear which raises question as to their stability under large noise.

%
% -------- related to event based cameras ---------------------

The amount of literature published on frequency detection with event based cameras is
quite limited. Among the earlier works is research on blinking LED marker
detection in Ref.~\cite{censi_strubel}.
The authors outline an algorithm to find which if any of a set of
frequencies is present in a signal. Their key idea of
detecting signal peaks via polarity transitions is what we refer to as
the baseline method in Sec.~\ref{sec:baseline}. In contrast to our work they know beforehand
what the frequency spectrum looks like, but then have to deal with the
thorny issue of interference between signals of multiple
frequencies. The authors use the transition from ON to OFF events
which we confirm to have less jitter than the transition from OFF to
ON events. Curiously this fact was overlooked by the later work of
Ref.~\cite{hoseini_linares} who perform frequency detection based on
the OFF to ON transition (their Fig.~4). Like in the present work,
Ref.~\cite{hoseini_linares} creates a frequency image of the spinning
propellers of a quad rotor, but unlike us they do not use a digital
filter and present an FPGA implementation whereas our algorithm runs
on a CPU. Except for some time averaging of the
estimated frequency and using the OFF/ON instead of the ON/OFF
transition, Ref.~\cite{hoseini_linares} algorithm in their Fig.~3 is
comparable in complexity to our baseline method (see
Sec.~\ref{sec:baseline}) and although their
work predates ours by four years we believe that with a carefully
optimized implementation this could have been done in software already
back then, in particular since their ATIS camera has a 304x240
resolution whereas our camera has 640x480 pixels.

Other related work on per-pixel temporal analysis for event
based cameras is focussed on filtering out periodic signals rather than
detecting them. Ref.~\cite{wang_yuan} develops a comb filter in a
continuous-time approach, but in this case the frequency is assumed to
be known, whereas our goal is to detect the frequency in the first place.

Similar in spirit to our work is the incremental Fourier analysis
of event based camera signals in Ref.~\cite{sabatier}. There the emphasis
too is on exploiting the neuromorphic, asynchronous nature of the
camera during frequency analysis. In contrast however, Ref.~\cite{sabatier}
performs a {\em spatial} frequency analysis whereas here the focus is
on {\em temporal} frequency detection. We could not find a way to
transfer their ideas efficiently from the equidistant spatial domain
to the irregularly spaced time domain.

The digital filter we propose in
Sec.~\ref{sec:digital_filter} and implement on the host
side operates in {\em event time}, meaning it does not utilize the wall
clock time stamps that the camera provides along with each event. Such
circuits are known~\cite{tsividis_tutorial} as asynchronous digital
signal processors and have been realized in hardware already in the
early 1990s~\cite{jacobs_digital_dsp}. This opens the possibility of
implementing our proposed filter in hardware close to the sensor
without adding the complexity of an additional clock.

\section{Method}
\label{sec:method}
%-------------------------------------------------------------------------
\subsection{Notation}
Following the convention of Ref.~\cite{gallego_delbrueck_orchard} a
pixel's brightness $L$ is defined to be the logarithm of its photo current
$I$:
\begin{equation}
  L = \log(I)\ .
  \label{eq:definition_brightness}
\end{equation}
Changes in brightness are signaled by events
$e_k = (t_k, p_k)$ where $k$ is the event index, $t_k$ the
time stamp, and $p_k \in \{-1, +1\}$ is the polarity of the event.

A brightness increase by more than a threshold $\con$ is
signaled with an ON event of polarity $p_k=+1$, whereas a brightness
decrease by more than  $\coff$ will result in an OFF event of
polarity $p_k=-1$:
\begin{equation}
  \Delta L_k = p_k \conoff\ .
  \label{eq:brightness_change}
\end{equation}

To simplify the notation the customary pixel location
$\mv{x}_k$ for the event is omitted since in this work no
spatial filtering is performed. One must bear in mind though that the
thresholds $\conoff$ vary from pixel to pixel
~\cite{wang_ng} similar to fixed pattern noise in a frame based camera.

\subsection{Baseline}
\label{sec:baseline}
\begin{figure}[t]
  \centering
  \includegraphics[width=1.0\linewidth]{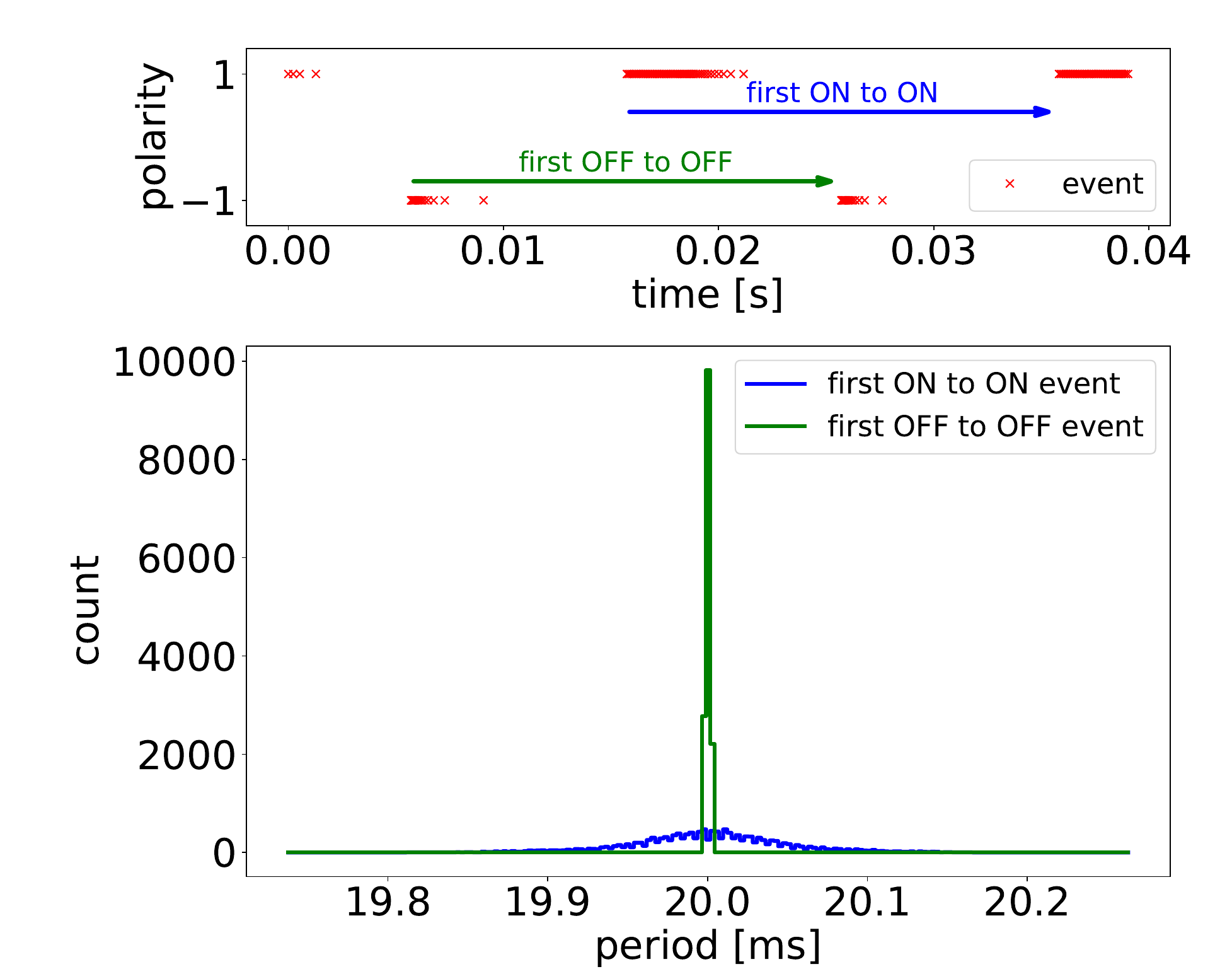}

  \caption{Frequency estimation with the baseline method for a 50Hz
    square wave signal. The top panel shows the raw events,
    and the two different ways to estimate the signal period. Note
    that although the brightness changes abruptly between up and down, the ON and
    OFF events spread out due to ``motion blur``, and much more so for
    ON events (see \cite{hu_liu}, section 3.1). The bottom panel shows
    the dramatically lower variance ($\sigma = 1.1\mathrm{us}$, green
    line) when
    using the ON/OFF transition for period estimation as opposed to
    OFF/ON ($\sigma=260\mathrm{us}$, blue line).}
\label{fig:baseline}
\end{figure}

For simple signals consisting of one monotonically rising and falling
section each per cycle the period of the signal can be found without
resorting to reconstructing the brightness~\cite{censi_strubel},
\cite{hoseini_linares}. As shown in Fig.~\ref{fig:baseline} (top
panel) for the
example of a 50Hz square wave signal it is sufficient to record the
elapsed time whenever the event polarity changes.

Measuring the signal period by detecting the time of successive OFF to
ON event transitions as proposed in Ref.~\cite{hoseini_linares} yields
much less accurate results than detecting the transition between
successive ON to OFF events. This was already noted in
Ref.\cite{censi_strubel} but is here quantitatively shown in the lower panel
of Fig.~\ref{fig:baseline} where the ON/OFF transition period has a
standard deviation of $\sigma = 1.1\mathrm{us}$ (the time stamp accuracy of the camera is
1us!), whereas using OFF/ON transitions yields
$\sigma=260\mathrm{us}$. The underlying reason for this is most likely
that before the first ON event the pixel's photo current has fallen to very low
levels which leads to long response times as documented for the DVS
sensor (\cite{hu_liu}, section 3.1).

Obviously the baseline algorithm is highly efficient and fully
asynchronous. We find it to be surprisingly robust when the
camera is used in its default bias configuration or when the
fundamental frequency of the signal is very high (c.f.~Sec.~\ref{sec:frequency_sweep}).
This is owed to the
low pass behavior of the sensor's photo receptor and source
follower~\cite{hu_liu} which largely filters out higher frequencies
such that when the remaining signal reaches the comparator it only
produces one ON sequence and one OFF sequence of events per cycle.

\subsection{Reconstruction by Digital Filtering}
\label{sec:digital_filter}
\begin{figure}[t]
  \centering
  \includegraphics[width=1.0\linewidth]{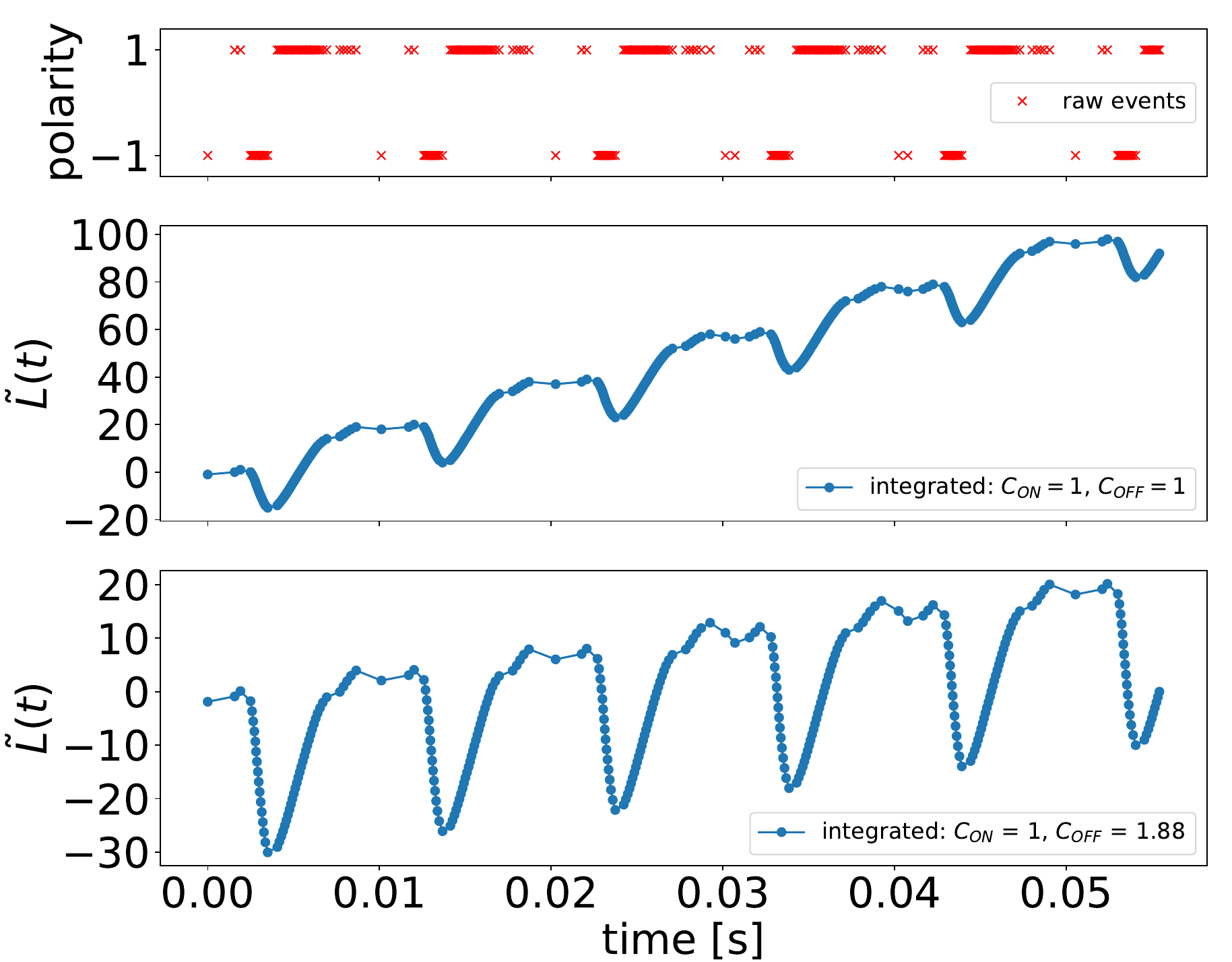}

  \caption{Multiple periods of a single pixel signal with fundamental
    frequency of 100hz. The top panel shows the raw signal, for which
    the baseline frquency detection fails. The middle panel displays
    the (non-stationary) naive illumination
    reconstruction (cumulative sum of polarities). The bottom panel
    shows the reconstruction obtained after removing the {\em global}
    (but not local!) trend by adjusting the OFF threshold to
    $\coff=1.88$. Note the DC component during the first cycle and the
    drift due to the $\con$/$\coff$ ratio varying with time.}
   \label{fig:simple_detrend}
\end{figure}

While the baseline method is simple and fast it fails for signals that
have more than one ON and OFF section per cycle such as the one
shown in Fig.~\ref{fig:simple_detrend}. Such signals are more likely
to be encountered when the sensor biases are tuned for low latency and the
sensor's region of interest (ROI) is reduced.

To address the baseline's shortcoming, the following approach is proposed:
\begin{itemize}
\item reconstruct the brightness approximately by means of a digital filter,
\item detect when the reconstructed signal crosses zero from above.
\end{itemize}
This method is significantly more robust to noise and will correctly
detect the fundamental period in many situations where the baseline
does not.

The simplest way to reconstruct brightness is to start
from Eqn.~(\ref{eq:brightness_change}) and to assume $\con = \coff = 1$:
\begin{equation}
  \ltil_{\mathrm{naive}}(t) = \sum_{k | t_k \leq t} p_k
  \label{eq:naive_reconstruction}
\end{equation}
As the middle panel of Fig.~\ref{fig:simple_detrend} shows, $\con =
\coff$ is a poor assumption that leads to drift in the
reconstruction. Not only are $\con$ and $\coff$ sensor
parameters that can be separately configured, but they also change with
the overall illumination level and from
pixel to pixel~\cite{lichtsteiner_posch_delbrueck}. Furthermore we
find the ratio of ON to OFF events to vary with the frequency of the
time-varying light signal (Fig.~\ref{fig:roi_reconstruction}) and
to fluctuate over time, rendering a calibration difficult. Said
fluctuations cause the reconstructed signal shown in the bottom panel of
Fig.~\ref{fig:simple_detrend} to trend during the time
period shown although the thresholds $\conoff$ have been adjusted to
remove any drift over the full duration of the sample. These
observations suggest reconstructing the brightness by performing local
detrending and integration first, followed by a suitable high pass
filter to remove any DC component (see again bottom panel of
Fig.~\ref{fig:simple_detrend}).

This is exactly what the proposed digital IIR filter does.
Before describing the filter however a brief digression
regarding sampling and time stamps is necessary.
Typically digital filters are based on data points obtained by
sampling at regular time intervals, but an event based camera
provides samples only when the brightness of a pixel changes by more
than $\conoff$. The digital filter presented here
takes as input directly the event polarity $p_k \in \{-1, +1\}$ of
event $k$. This also constitutes uniform sampling but in
{\em event time}, not in {\em wall clock time}. Thus the filter 
ignores all time stamps $t_k$. In fact time stamps are only taken into
consideration later when zero-level 
crossings are detected (Sec.~\ref{sec:zero_level_crossings}).
Our event time based method is in contrast
to the filtering performed in \cite{wang_yuan} which operates in
continous time. Since the filter does not make use of any time stamps
its output is invariant to the (wall clock) time scale, i.e. the
frequency of the signal.

As motivated earlier, the first stage of the filter will locally detrend the input
signal. This is equivalent to subtracting the moving average polarity
$\bar{p}$ from the polarity $p_k$ itself:
\begin{equation}
  \label{eq:remove_average}
  \Delta \ldet_k := p_k - \bar{p}_{k-1}
\end{equation}
where $\bar{p}_{k}$ is the
moving exponential average with a mixing coefficient $\alpha \in [0, 1]$:
\begin{equation}
  \label{eq:exp_average}
  \bar{p}_k := \alpha\ \bar{p}_{k-1} + (1 - \alpha)\ p_{k}\ .
\end{equation}
In (\ref{eq:remove_average}) we have set $\con=1$ and $\coff=1$ because
the absolute scale of the signal does not matter for detecting
zero-level crossings and the averaging in (\ref{eq:exp_average}) will
compensate for the $\con/\coff$ ratio being different from unity.
The output of the first filter stage is the detrended change in brightness
$\Delta\ldet$, where the tilde indicates that this
is an approximate reconstruction.

Next, the detrended changes in brightness are integrated up:
\begin{equation}
  \ldet_k = \ldet_{k - 1} + \Delta\ldet_k
  \label{eq:integ_bright}
\end{equation}
and passed through a high-pass filter to remove the DC component and
remaining low-frequency noise as discussed earlier:
\begin{align}
  \ltil_k &= \beta \ltil_{k-1} + \frac{1}{2}(1 + \beta)(\ldet_k - \ldet_{k-1})   \label{eq:high_pass_1}
\\
  & = \beta \ltil_{k-1} + \frac{1}{2}(1 + \beta)\Delta\ldet_k \label{eq:high_pass_2}\ .
\end{align}
    
Evidently the integration in (\ref{eq:integ_bright}) can be skipped
since the high-pass filter (\ref{eq:high_pass_1}) immediately takes
differences of its input. The particular way the filter parameter
$\beta\in[0, 1]$
is introduced in (\ref{eq:high_pass_1}) ensures that the high pass has unit gain at the Nyquist
frequency $\omega = \pi$.

\begin{figure}[t]
  \centering
  \includegraphics[width=1.0\linewidth]{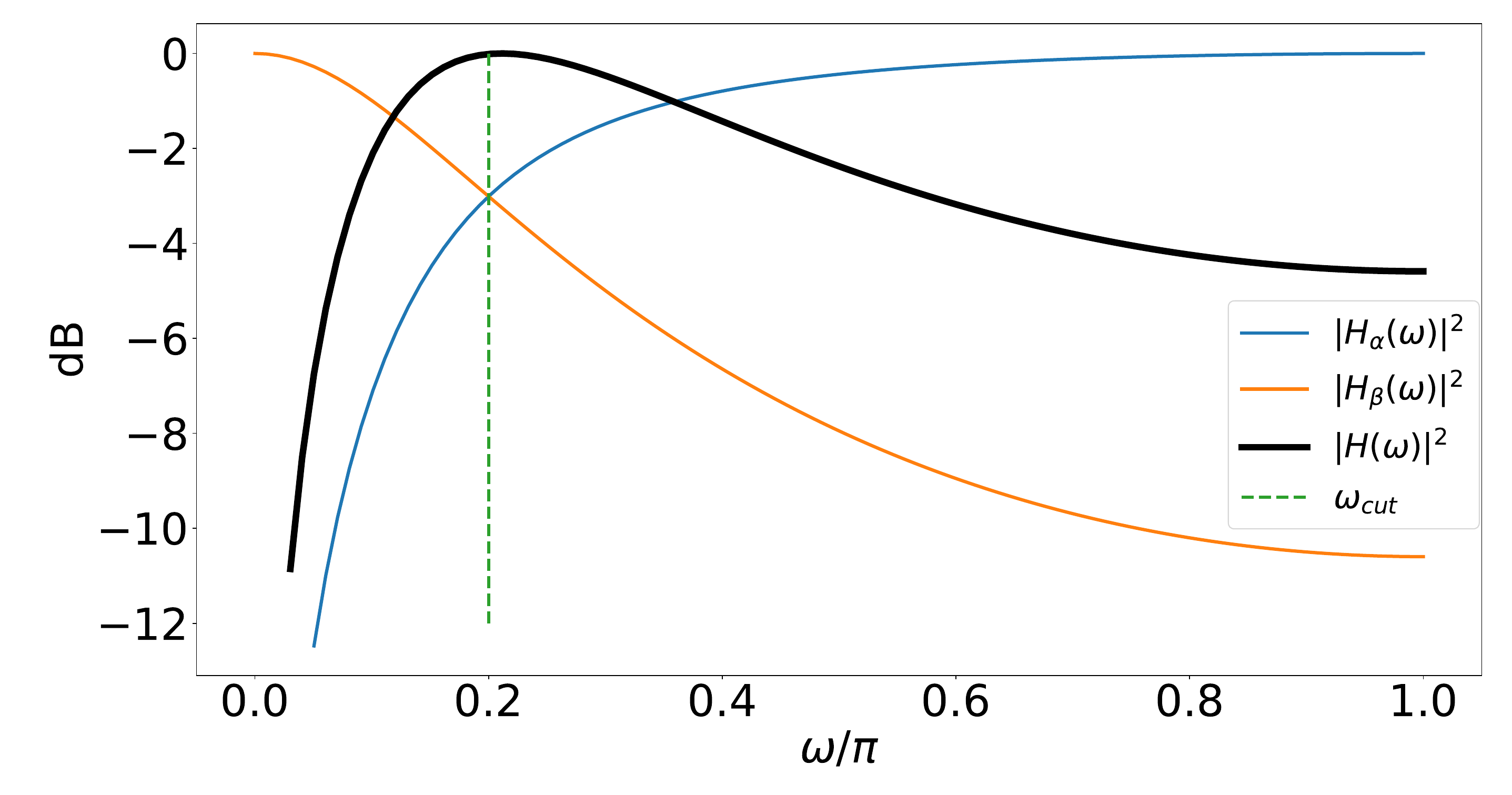}
  \caption{Bode magnitude plot of the transfer functions of Eq.(\ref{eq:transfer_functions}),
    but normalized to unity at their respective maxima. The coefficients $\alpha=0.51$ and
    $\beta=0.54$ have been set according to (\ref{eq:alpha_cut}), (\ref{eq:beta_cut}) to result in a identical cutoff (-3dB) frequency of
    $\ocut=0.2 \pi$ for both high and low pass, resulting in a bandpass center
    frequency close to $\ocut$.}
   \label{fig:transfer_functions}
\end{figure}

Combining (\ref{eq:remove_average}), (\ref{eq:exp_average}) and
(\ref{eq:high_pass_2}) and utilizing the $z$-transform
(\ref{eq:z_transform}) yields the simple and efficient second
order recursive filter equation for the approximate reconstructed
brightness:
\small
\begin{equation}
  \label{eq:filter_recursion_ab}
  \ltil_k = (\alpha + \beta)\ltil_{k-1} - \alpha\beta\ltil_{k-2} +
  \frac{\alpha}{2}(1 + \beta)(p_k - p_{k-1})\ .
\end{equation}
\normalsize
How should the filter coefficients $\alpha$ and $\beta$ of
Eq.~(\ref{eq:filter_recursion_ab}) be set?
Insight can be gained by applying the $z$-transform well known from the digital signal
processing literature~\cite{proakis_manolakis} to (\ref{eq:filter_recursion_ab}):
\begin{equation}
  \label{eq:z_transform}
  Y(z) = \sum_{n=-\infty}^{\infty}y_nz^{-n}\ .
\end{equation}
Under this transform Eqn. (\ref{eq:filter_recursion_ab}) becomes:
\begin{equation}
  \label{eq:transfer_functions}
  \begin{split}
    \ltil(z) &= H_\alpha(z) H_\beta(z) P(z)\\
    H_\alpha(z) &= \left(\frac{\alpha(z - 1)}{z - \alpha}\right)\\
    H_\beta(z) &= \frac{1}{2}\frac{z(1 + \beta)}{z - \beta}\ .
  \end{split}
\end{equation}
The transfer function $H(z) = H_\alpha(z) H_\beta(z)$ connecting input
polarities $P(z)$ to reconstructed brightness $\ltil(z)$ is the product
of a high pass $H_\alpha(z)$ and a low pass $H_\beta(z)$ component
which together form a band pass, see
Fig.~\ref{fig:transfer_functions}.

A few words are in order as to why $H_\alpha(z)$ acts as high
pass and $H_\beta(z)$ as a low pass. Although the averaging procedure in (\ref{eq:exp_average})
suggests that $H_\alpha$ may be a low pass, it is in fact a high pass
due to the first term on the r.h.s. of (\ref{eq:remove_average}) being
an all pass from which then the low pass is subtracted. Likewise, despite
$H_\beta$ involving the high pass filter (\ref{eq:high_pass_1}) it also contains the
integration (\ref{eq:integ_bright}) which is intrinsically a low pass
operation, resulting in $H_\beta$ actually having low pass
characteristics.

Since each sample corresponds to a single event, it is more intuitive
to think of the filter in terms of the cut-off period $\Tcut = 2\pi/\ocut$ rather than the
cutoff frequency $\ocut$, and to consider a signal to consist of
$\nperiod=\non+\noff$ events. To preserve the fundamental frequency of the original signal one
should first set $\alpha$ such that the cutoff period $\Tcut$ for the
$H_\alpha$ high pass is larger than $\nperiod$. The cutoff
frequency of $H_\alpha$ depends on $\alpha$, and is defined to be the
frequency where $|H_\alpha|^2$ has dropped to half its
maximum. Substituting $z=\exp(j\omega)$ into
(\ref{eq:transfer_functions}) yields the relationship between cutoff
frequency and $\alpha$:
\begin{equation}
  \label{eq:alpha_cut}
  \alpha = \frac{1 - \sin(\ocut)}{\cos(\ocut)}\ .
\end{equation}
An analogous procedure can be used to find the relationship between
$\beta$ and the cutoff frequency for the low-pass $H_\beta$, which
again is where the signal power is attenuated by half:
\begin{equation}
  \label{eq:beta_cut}
  \beta = (2-\cos(\ocut)) - \sqrt{(2 - \cos(\ocut))^2 - 1}\ .
\end{equation}
Similar design considerations as for $\alpha$ apply to the choice of $\beta$: the
cutoff frequency of the low pass filter $H_\beta$ should be high enough to just
let the signal pass while not being any higher to avoid damping
the summation process in (\ref{eq:high_pass_2}) that is crucial to the
brightness reconstruction. A reasonable choice is then setting the
cutoff frequency for $H_\alpha$ and $H_\beta$ to be equal, i.e. a
single cutoff period $\Tcut$ is chosen for both high and low pass
filters. Since $\ocut$ is usually well smaller than one and since both
(\ref{eq:alpha_cut}) and (\ref{eq:beta_cut}) happen to have the same
Taylor series expansion to second order
$\alpha = 1-\ocut + \frac{1}{2}\ocut^2 + \mathcal{O}(\ocut^{3})$ in practice
this implies $\alpha = \beta$ such that $H$ has
a double pole on the real axis at $z=\alpha$.

\begin{figure}[t]
% for example we are using the wiggle_100hz bag which has 845.1 cycles
% with 14262 OFF (16.9/cycle) and 30230 ON (35.8/cycle) events, or
% 42.7 events/cycle total
% The target cutoff period is then 2 * (2*36) = 144
  \centering
  \includegraphics[width=1.0\linewidth]{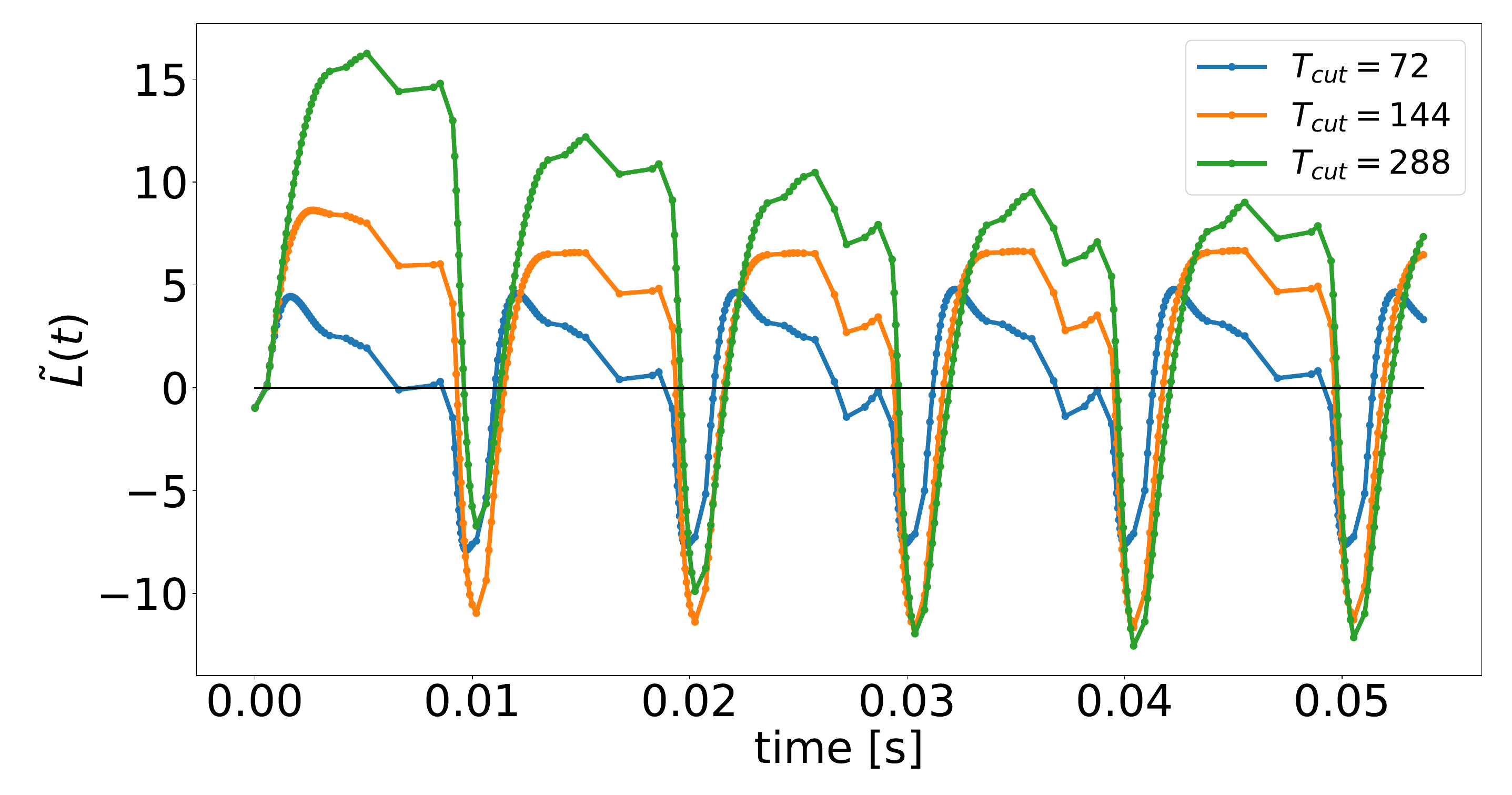}
  
  \caption{Single-pixel reconstructed brightness $\ltil(t)$ when the
    filter (\ref{eq:filter_recursion_ab}) is applied with different
    cutoff periods $\Tcut$ to the signal shown in Fig.~\ref{fig:simple_detrend}.
    The orange line with $\Tcut=144$ corresponds to the recommended setting according to
    (\ref{eq:setting_tcut}). Notice how the DC component is removed
    faster than for $\Tcut=288$ (green line) while still providing a
    sufficiently high quality reconstruction. The blue line
    ($\Tcut=72$) shows how
    the reconstruction suffers when setting the cutoff period too
    small, eliminating low frequencies too aggressively.}
   \label{fig:filter_examples}
\end{figure}

Fig.~\ref{fig:filter_examples} shows the result of filtering the signal
from Fig.~\ref{fig:simple_detrend} with different cutoff periods. To achieve
good reconstruction both ON and OFF event sections must be
reconstructed properly so in practice a conservative choice is
keeping $\Tcut$ a factor of two larger than the signal period implied by the
larger of $2 \noff$ and $2 \non$:
\begin{equation}
  \Tcut = 4 \max(\noff, \non)\ .
  \label{eq:setting_tcut}
\end{equation}
The orange line in Fig.~\ref{fig:filter_examples} demonstrates that picking
$\Tcut$ in this way is indeed a good tradeoff between accurately
reconstructing the original signal and filtering out low-frequency components.

\subsection{Finding Zero Level Crossings}
\label{sec:zero_level_crossings}
\begin{figure}[t]
  \centering
  \includegraphics[width=1.0\linewidth]{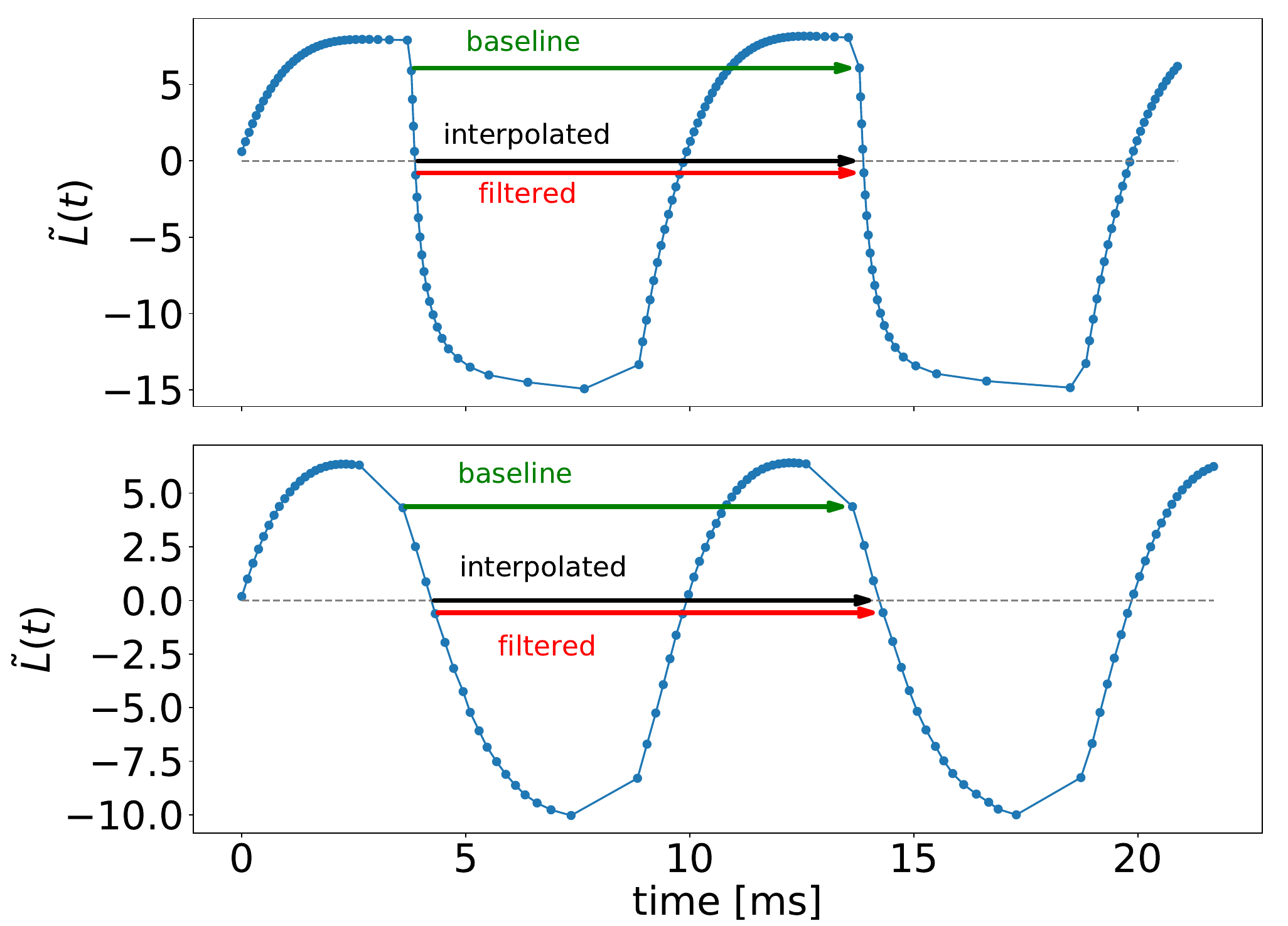}
  \caption{Three different ways of measuring the signal period:
    1) baseline (green) measuring the time between successive ON to
    OFF transitions, using the time stamp of the first OFF event,
    2) filtered (red) taking the time stamp of the first event after
    zero level crossing from above, 3) interpolated (black), taking
    the straight-line interpolated time between the events before and
    after the zero level crossing. The top panel shows the
    reconstruction for a square wave, the bottom for a triangle wave,
    both at 100Hz.}
   \label{fig:reconstruction_square_triangle}
\end{figure}

Once the approximate reconstruction is complete the zero level
crossings can be detected and the signal period derived from that.
As Fig.~\ref{fig:baseline} clearly shows, for the baseline method the
ON to OFF transition gives much more accurate period estimates than the OFF to ON transition.
This carries over to the reconstructed signal as well so we only
consider the level crossing {from above to below zero} when measuring
the period. One can then assign the first event time
stamp after the level crossing to be the time of the level
crossing or linearly interpolate between the two events straddling the
zero crossing. Such interpolation schemes are commonly used for power
line frequency estimation \cite{mendonca_power_grid} where the
signal has low noise but high accuration estimates are
desired. Figure~\ref{fig:reconstruction_square_triangle} visualizes 
these two different methods of measuring the period along
with the baseline approach. The top panel
shows the reconstruction of a square wave, the bottom of a triangle
wave where the logarithm of the current going through the
illuminating LEDs was ramped up and down linearly.

\subsection{Dark Noise Filter}
\label{sec:dark_noise_filter}
\begin{figure}[t]
  \centering
  \includegraphics[width=1.0\linewidth]{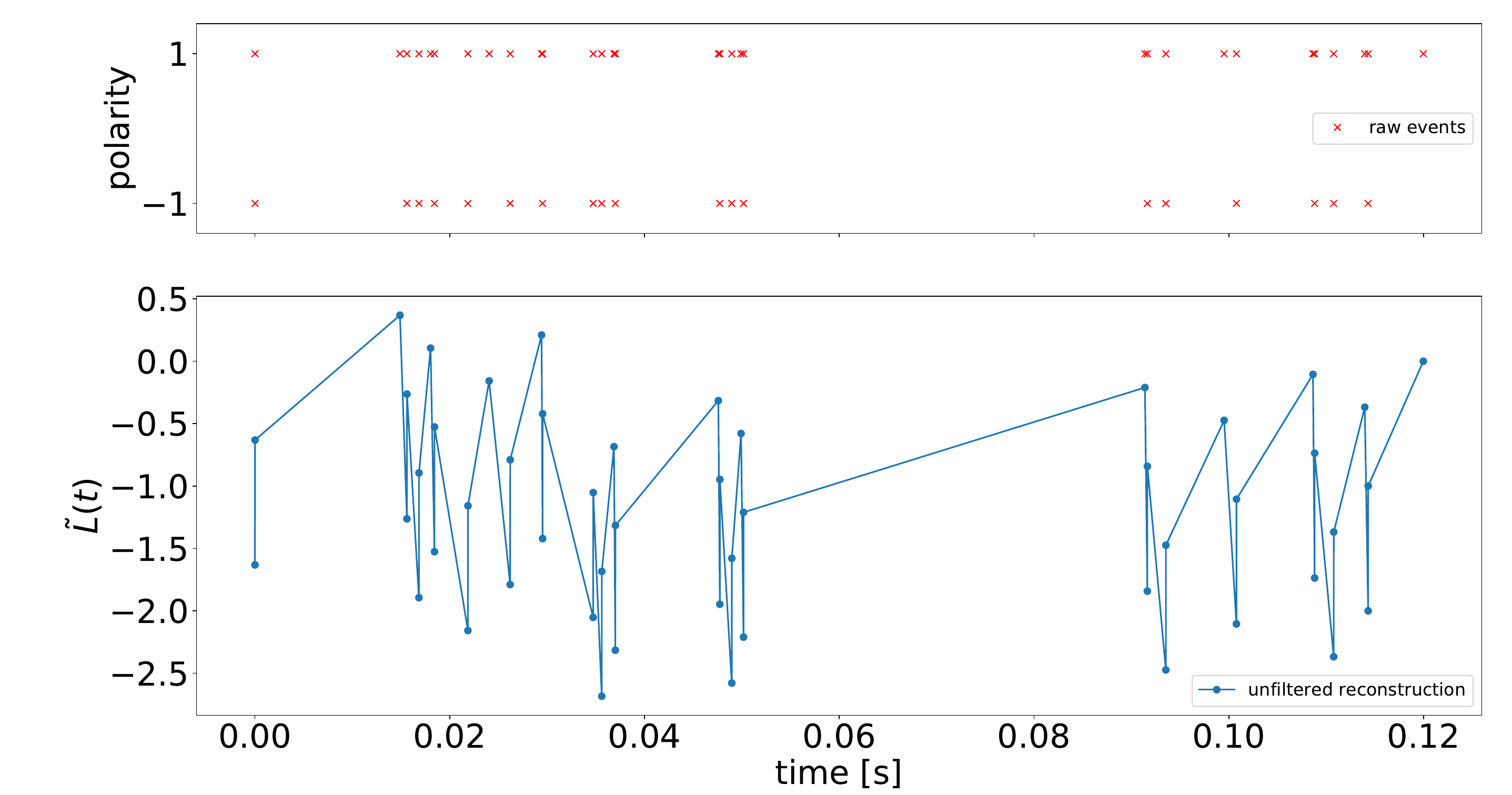}
  
  \caption{When tuned to ``fast (single pixel)'' bias settings
    (Tab.~\ref{tab:bias_settings}) the sensor emits dark 
    noise events with a distinct temporal signature: first an event occurs
    followed after some time by a pair of OFF/ON events that
    occur in rapid succession, typically within less than 15us. The
    top panel shows the raw noise event polarities, the bottom the approximate
    reconstruction. Such noise events thwart any frequency analysis and
    need to be filtered.}
   \label{fig:dark_noise}
\end{figure}

When operating the camera under ``fast (single pixel)'' bias settings (see
Tab.~\ref{tab:bias_settings}) the sensor emits
a large number of noise events when the illumination is low. Fig.~\ref{fig:dark_noise} shows a sample collected in complete darkness.
Fortunately the dark noise events have a very characteristic
temporal signature:  first an event happens (often ON, but could be either polarity)
followed after some
time by a pair of OFF/ON events that occur in rapid succession,
typically within less than 15us. Hence the remedy is to remove any
triplet of events where the current event has polarity ON, the
previous event occured within less than 15us and has polarity OFF, and
the event before that one occured a significant time earlier, e.g. by more
than 15us. The latter test will avoid filtering out genuine high frequency
signals. Since the spacing in arrival times of the noise OFF/ON events
are found to approximately follow an exponential distribution with a
mean arrival period that is much larger than 15us, only a small number of noise
events will slip through the filter. Note that this dark noise filter
utilizes timestamps and is completely unrelated and separate from the
digital filter presented in Sec.~\ref{sec:digital_filter}.

\begin{figure}[t]
  \centering
  \includegraphics[width=1.0\linewidth]{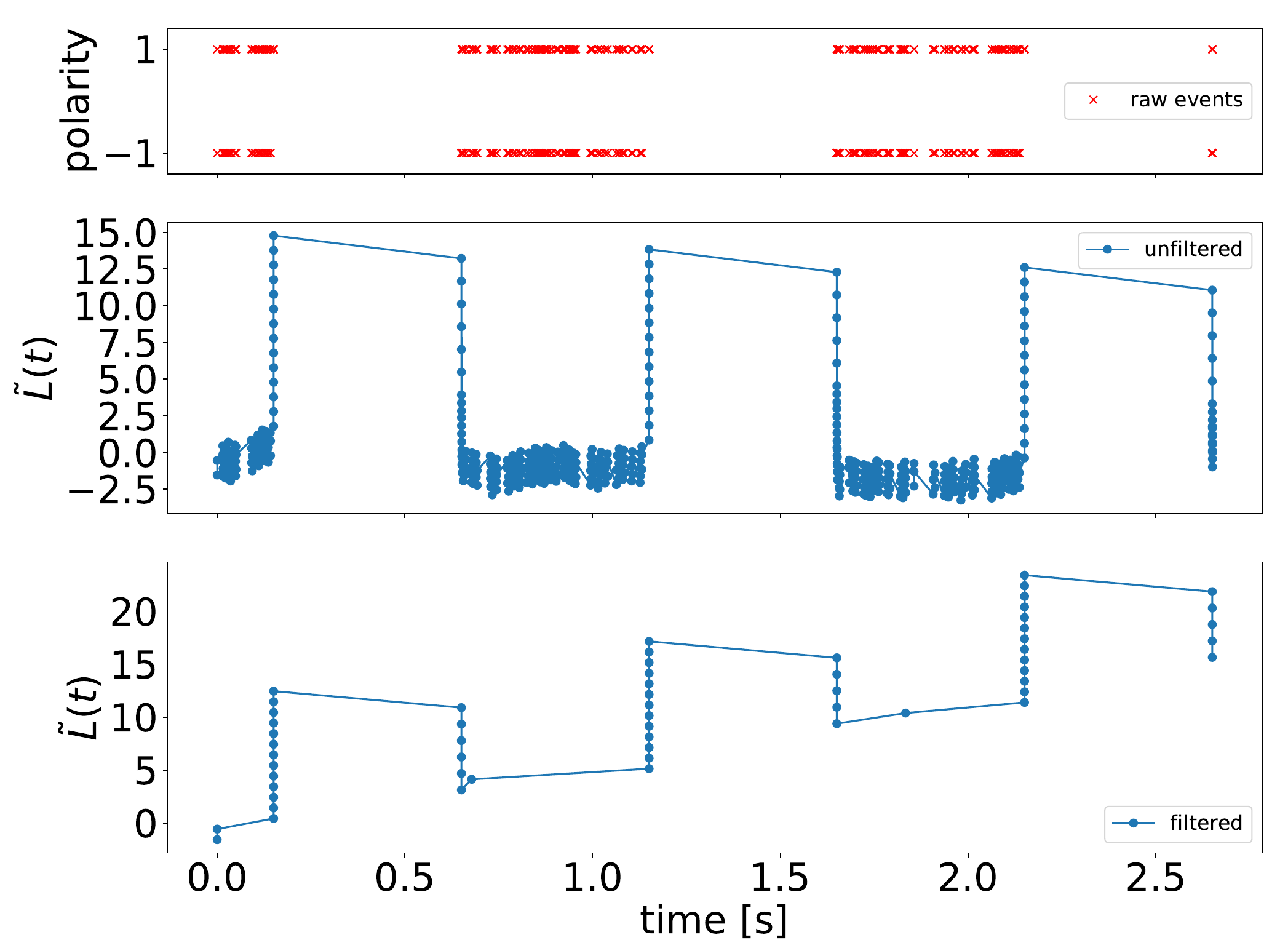}
  \caption{Approximate reconstruction of a 1 Hz square wave signal
    with (bottom panel) and without (middle panel) dark noise
    filtering.}
   \label{fig:dark_noise_filtered}
\end{figure}

Noise filtering is particularly beneficial for low frequency signal
detection as demonstrated in Fig.~\ref{fig:dark_noise_filtered}. Note that
the dark noise filtering alters the ratio of ON to OFF events
because more noise events occur when the light is being switched
off. However the digital filter described in
Sec.~\ref{sec:digital_filter} will readily compensate for
such an imbalance so long as it does not change abruptly with time.

While the noise reduction in Fig.~\ref{fig:dark_noise_filtered} looks
quite impressive, some qualifying statements are necessary.
\begin{enumerate}
  \item The noise arises only when the sensor bias settings are tuned
    for high speed operation on a single pixel, {\em not} for
    reasonable full-sensor bias settings.
  \item It shows the greatest benefits for detecting low frequency
      signals, but for such signals there is no reason to tune the
      sensor biases for high speed. One can however argue that the
      dark noise filter does increase the captured frequency range of
      the sensor without requiring changes to the bias settings.
  \item Since the filter does not produce
    output until three subsequent events have occured it introduces a
    lag of potentially unbounded length in wall clock time. This can
    be difficult to deal with when requiring to synchronize on wall clock
    time during readout, i.e. when transitioning from neuromorphic to
    von Neumann computing.
\end{enumerate}

\section{Experiments}
In this section we will show how the methods presented in
Sec.~\ref{sec:method} work in practice and what their strengths and
weaknesses are. We also use the opportunity to showcase the practical
difficulties encountered regarding readout saturation and
lens flare.

All experiments are performed with a CenturyArks SilkyEvCam model
EvC3A, employing a third-generation Prophesee sensor with VGA 640x480
resolution and a maximum bandwidth of 50 million events/sec
(Mevs). For the single pixel datasets and the guitar frequency imaging
a Kowa LM35HC 35mm lens is used. All other recordings are obtained
with a Computar M0814-MP2 lens. Data is collected and processed on an HP Omen 15 (AMD Ryzen 7
4800H 2.9GHz) laptop with a ROS2 driver building on the Metavision 2.3.2 SDK.

To reduce noise level and avoid camera bandwidth
saturation the default programmable bias settings of the SilkyEVCam
tune the front end source follower to act as a low pass with a fairly
low cutoff frequency, making it difficult to analyze signals above
about 1kHz. For this reason different bias settings are used to
extend the usable frequency range of the camera as listed in
Tab.~\ref{tab:bias_settings}.

\subsection{Single Pixel Experiments}
\label{sec:single_pixel_experiments}
In this section a series of different experiments are presented that
show how the filter performs versus the baseline. To avoid sensor
readout bandwidth saturation the ROI hardware feature
of the sensor is used to restrict the activity to only the
center pixel ($x=319$, $y=239$). The camera is pointed at a white surface that is
illuminated by six LEDs positioned behind the camera and controlled by
a MOSFET based switch which is in turn controlled by a Teensy 4.1
board running at 600MHz CPU frequency. The illuminated surface, LEDs,
and lens are enclosed in a lightproof box to ensure that no stray light
affects the experiments.

To facilitate comparison the test signals generated with the Teensy
board are deliberately picked to be simple enough so the baseline
algorithm works, i.e. signals such as those shown in Fig.~\ref{fig:simple_detrend} are
avoided. For the square and triangle wave signals (Sec.~\ref{sec:square_wave})
the default bias settings (c.f.~Tab.~\ref{tab:bias_settings})
are sufficient.

\subsubsection{Square Wave}
\label{sec:square_wave}
\begin{figure}[t]
  \centering
  \includegraphics[width=1.0\linewidth]{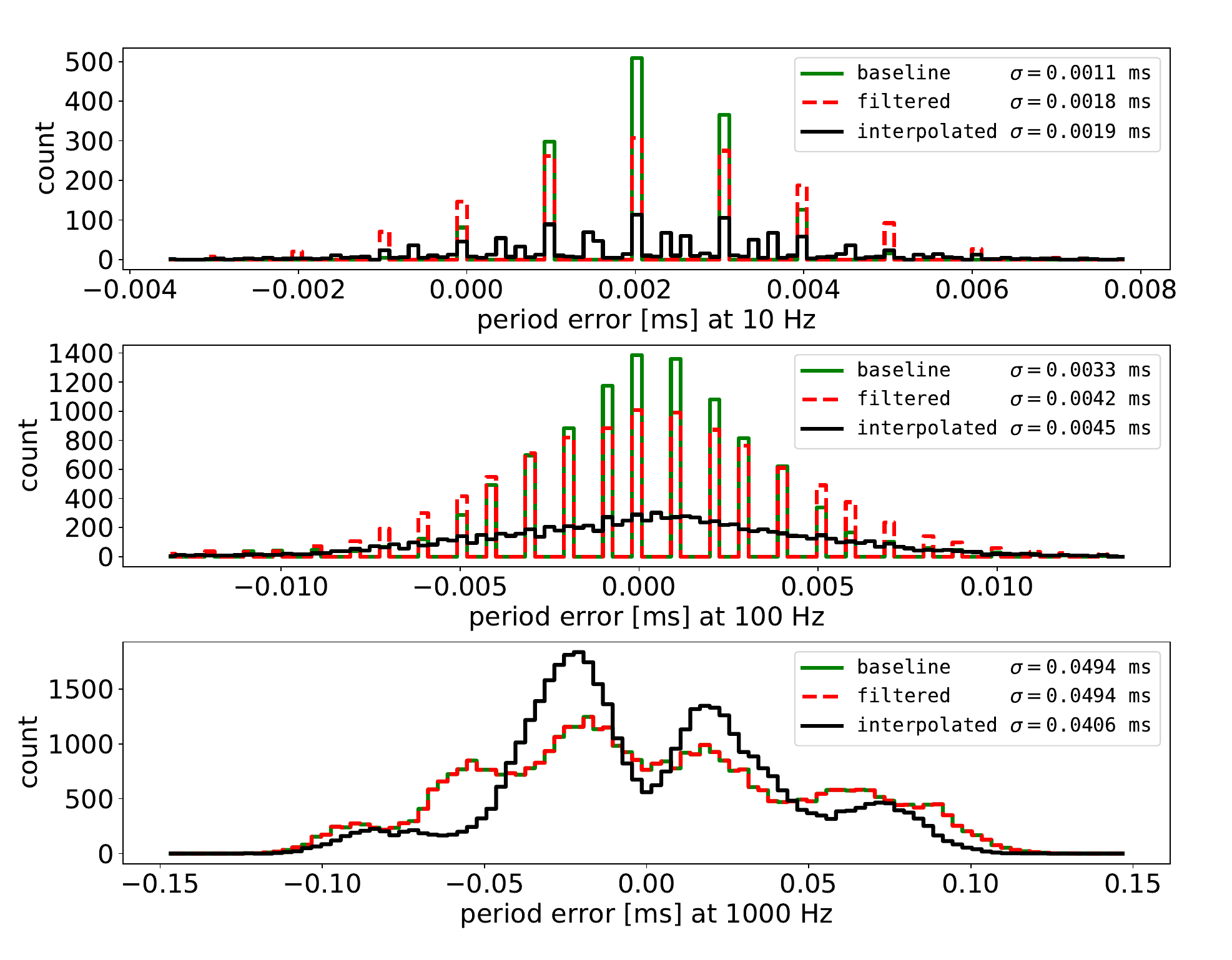}
  
  \caption{Signal period estimates for 10Hz (top panel, with $\Tcut=200$), 100Hz (middle
    panel with $\Tcut=190$) and 1kHz signals (bottom
    panel, $\Tcut=23$) using the baseline method
    (green), filtering (red), and filtering combined with
    interpolation (black). The $y$ axis gives the
    number of times a given period was detected, while the $x$ axis
    gives the sample's deviation from the ground truth period. Note
    that the accuracy of the camera time stamps is
    1us which is visible at 10Hz and 100Hz but not at 1kHz where the ON
    events are no longer separated in time from the OFF events due to
    the finite sensor front end bandwidth. This drives up the jitter
    and favors interpolation.}
   \label{fig:square_wave}
\end{figure}

\begin{figure}[t]
  \centering
  \includegraphics[width=1.0\linewidth]{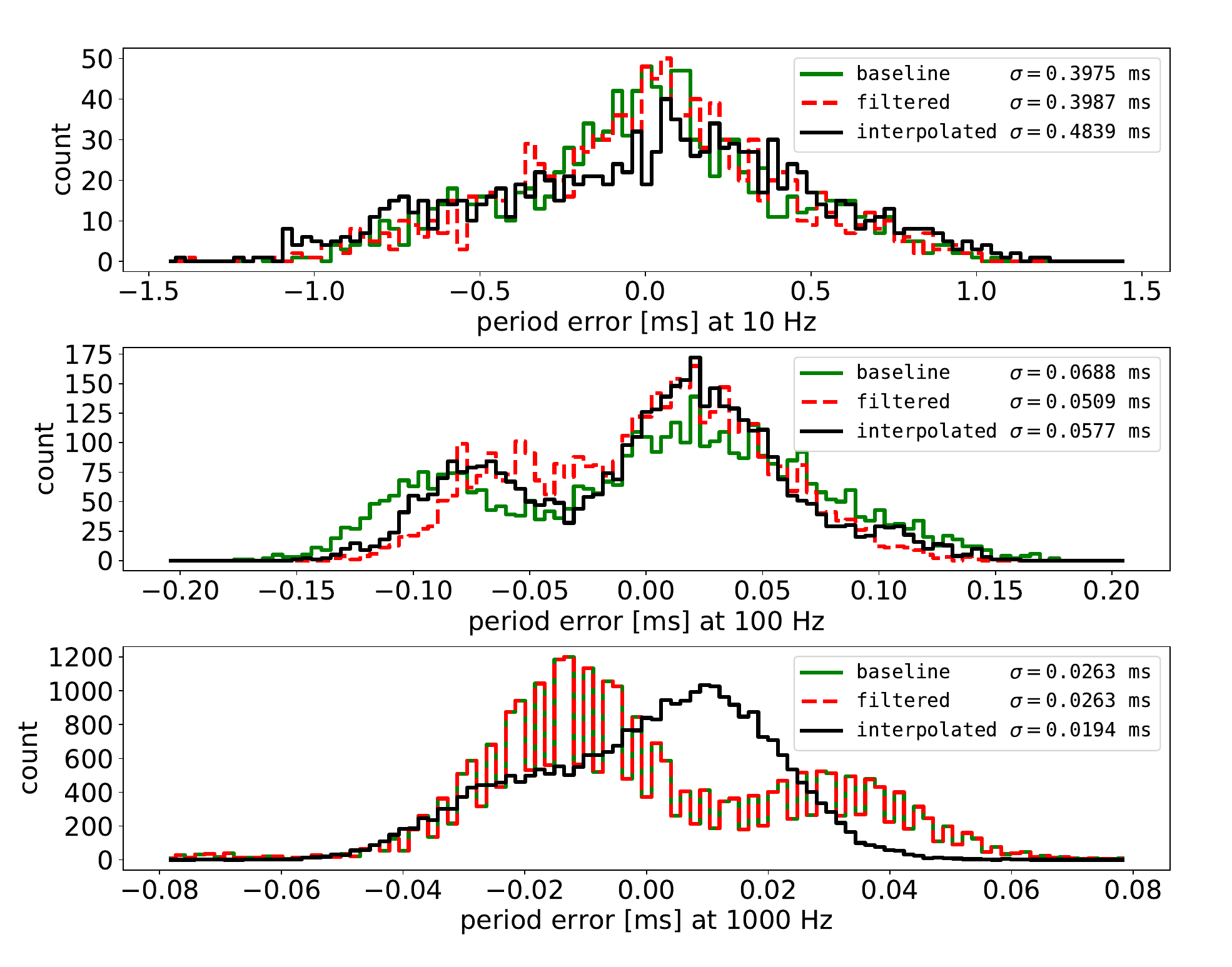}
  
  \caption{Signal period estimates for 10Hz (top panel, with cutoff
    period $\Tcut=158$), 100Hz (middle
    panel, $\Tcut=124$) and 1kHz signals (bottom
    panel, $\Tcut=17$) using the baseline method
    (green), filtering (red), and filtering combined with
    interpolation (black). The $y$ axis gives the
    number of times a given period was detected, while the $x$ axis
    gives the sample's deviation from the ground truth period. Due to
    the lack of a sharp signal onset the jitter is much larger than
    for the square wave (Fig.~\ref{fig:square_wave}), rendering
    reconstruction and interpolation beneficial already at 100hz.}
   \label{fig:triangle_wave}
\end{figure}

The first test signal is a square wave, i.e. the illuminating LED is
abruptly switched on and off. The signal's reconstruction is shown (at 100Hz) in
the top panel of Fig.~\ref{fig:reconstruction_square_triangle}. Notice that
although the change in illumination is abrupt, the events spread out
over almost the entire signal period. Reconstruction and signal period
measurements are performed for test frequencies of 10Hz, 100Hz, and 1000Hz (see
Fig.~\ref{fig:square_wave}). The camera
is operated with default bias parameters for which the sensor front
end starts attenuating the signal strongly already at 1kHz, reducing the number
of ON events per cycle from 50 (at 10Hz) to 48 (at 100Hz) to about 6
(at 1kHz). To within 0.1us all three methods arrive at the same estimates for the
mean: 100.002ms, 10.0002ms, and 1.0000ms. What causes the small over
estimation of the period is not clear. The jitter is also quite
similar as shown in Fig.~\ref{fig:square_wave}. For 10Hz and 100Hz the
baseline has the smallest jitter because the measurement start is
triggered only by the very first OFF event of the cycle which occurs
near maximum illumination and happens exactly when the light is
switched off. For 1kHz this is no longer the case due to the finite
bandwidth of the sensor frontend. The jitter increases substantially and
interpolation is now favored because it relies on two events (one
before, one after the zero level crossing).

\subsubsection{Triangle Wave}
\label{sec:triangle_wave}
The experiments in section~\ref{sec:square_wave} are now repeated for
a triangle wave for which the reconstruction is shown in the lower
panel of Fig.~\ref{fig:reconstruction_square_triangle}. Via pulse
width modulation at 146kHz - well above the sensor's front end low pass
 frequency - the controller ramps the logarithm of the illuminating 
LED current up and down linearly. Since in contrast to the square
wave the light is not switched off suddenly the occurence of the first OFF
event is less well localized in time and consequently the jitter is
much higher (compare to Fig.~\ref{fig:square_wave}), rendering
reconstruction and interpolation beneficial already at 100Hz. Like for the square wave
the estimated period means are very similar for all three
methods: 100.0026ms to 100.0029ms for the 10Hz signal, 10.0002ms (to
within 0.1us) for 100Hz, and 1.0000ms (to within 0.1us) for 1kHz.

\subsubsection{Frequency sweep}
\label{sec:frequency_sweep}
By default the SilkyEVCam used here boots up with bias
parameters that are tuned for full-sensor operation, where excessive
noise events can easily saturate the sensor readout bandwidth. Such
conservative settings don't allow for testing the limits of camera and
frequency detection alogrithm. Therefore the camera is now tuned for
speed with the ``fast (single pixel)'' settings as listed in
Tab.~\ref{tab:bias_settings}~\cite{metavision_bias_tuning}, and the
ROI is again restricted to the center pixel.
\begin{table}
\resizebox{\linewidth}{!}{
  \begin{tabular}{l|c|c|c|c|c|c|c}
    mode & diff & diff\_off & diff\_on & bias\_fo & bias\_hpf & bias\_pr& bias\_refr\\
    \hline
    default               & 299 & 221 & 384 & 1477 & 1448 & 1250 & 1500\\
    fast (LEDs)& 299 & 221 & 384 & 1430 & 1448 & 1250 & 1400\\
    fast (quad rotor)& 299 & 221 & 384 & 1399 & 1250 & 1250 & 1500\\
    fast (single pixel)& 299 & 234 & 374 & 1250 & 1525 & 1247 & 1300
  \end{tabular}
  }
  \caption{Bias settings of the SilkyEVCam camera with the Prophesee
    third-generation VGA sensor for the various experiments.}
  \label{tab:bias_settings}
\end{table}

With these camera settings the dark noise filter described in
Sec.~\ref{sec:dark_noise_filter} is essential for recovering the signal frequency.
The Teensy controller is programmed to sweep through the
frequencies exponentially from 1 Hz to 125 kHz and back by first
doubling, then halfing the frequency every 10 cycles.

\begin{figure}[t]
  \centering
  \includegraphics[width=1.0\linewidth]{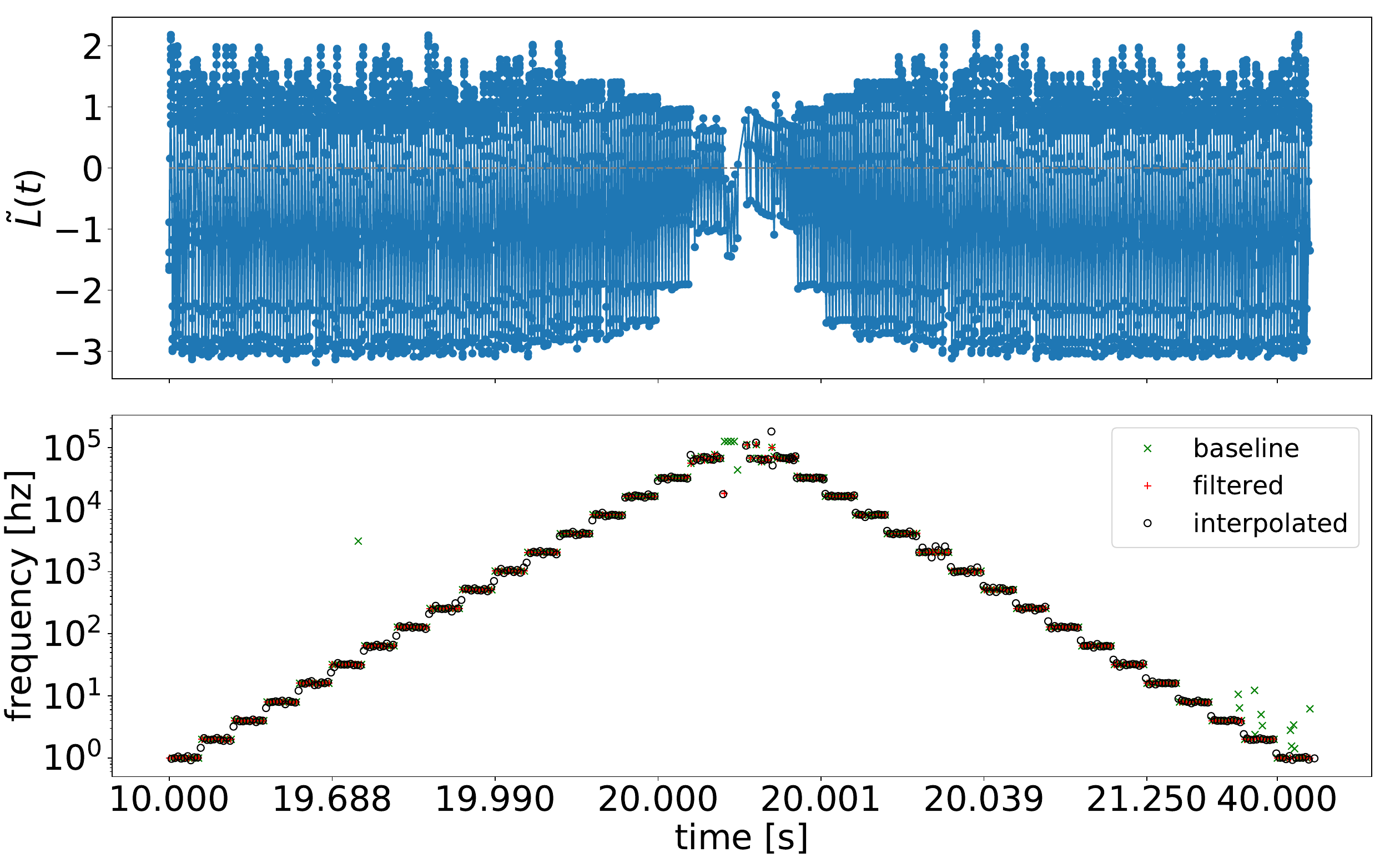}
  \caption{Exponential sweep from 1 Hz to 125 kHz and back to 1 Hz,
    with 10 cycles for each frequency. The top shows the approximate
    reconstruction of the signal ($\Tcut=25$), the bottom the
    baseline, filtered, and interpolated frequencies. Note how the
    time on the $x$ axis is warped such that the spacing of frequency
    detections is uniform. The top panel shows how the amplitude of
    the signal i.e. the number of events per cycle drops off rapidly
    for frequencies above about 10kHz. The bottom panel shows that
    both baseline and filtering algorithm work well over a wide range
    of frequencies, with filtering being more robust at low
    frequencies and the baseline performing better at maximum
    frequency (cf. Fig.~\ref{fig:frequency_sweep_detail})}
  \label{fig:frequency_sweep_overview}
\end{figure}

\begin{figure}[t]
  \centering
  \includegraphics[width=1.0\linewidth]{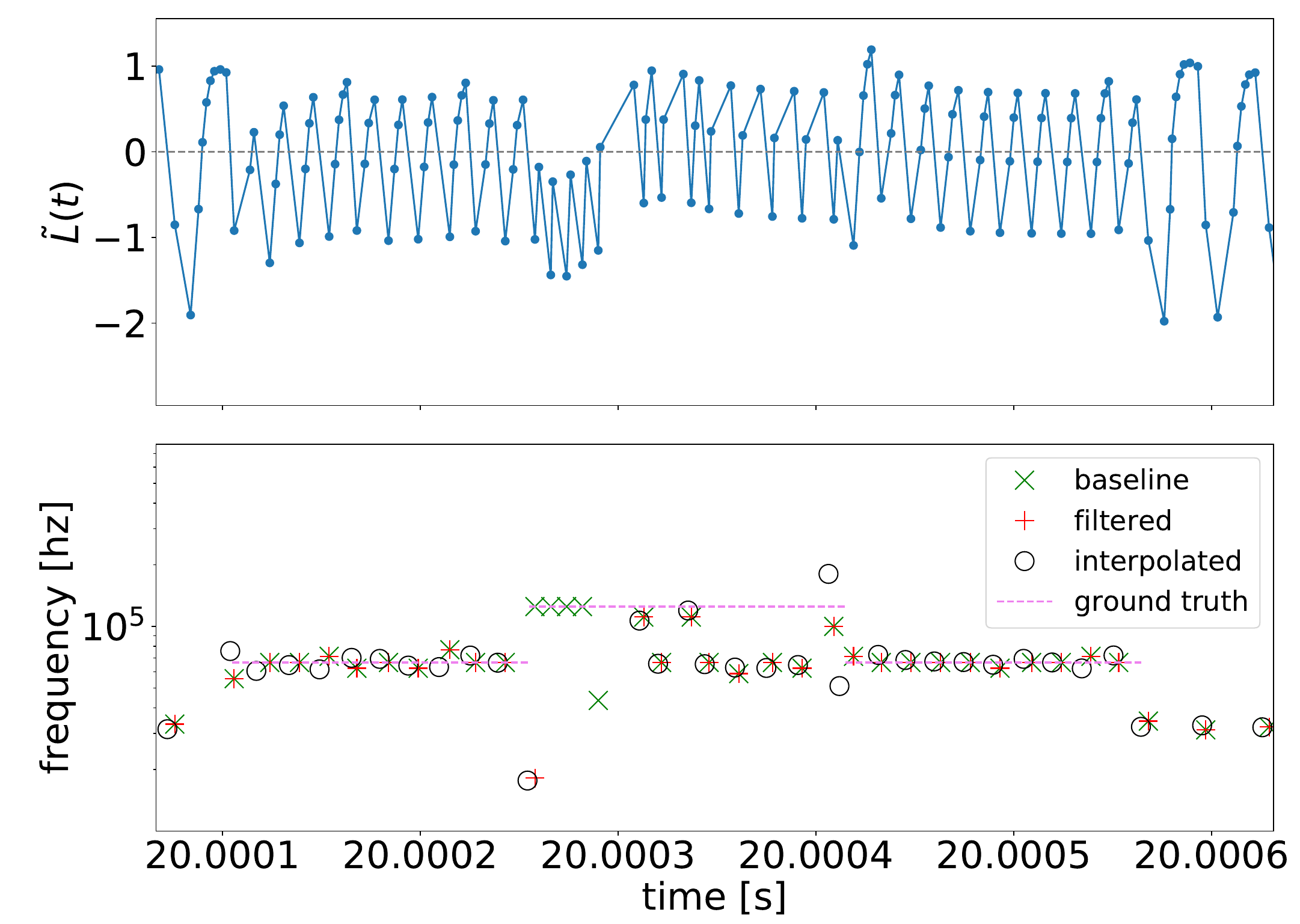}
  \caption{Zoomed in view of Fig~\ref{fig:frequency_sweep_overview}, but now {\em without} warping of the time axis. Shown are the last and first few cycles of the signal at 32 kHz, then in the middle 67 kHz and 125 kHz for which dashed ground truth lines are drawn. At 125 kHz the camera is at its limits and can no longer deliver a reliable signal. The first four cycles are correct, then OFF events are missing intermittently. The baseline method correctly detects the frequency for the first four cycles whereas the digital filter cannot adapt fast enough to the large change in the ratio of ON to OFF event.}
   \label{fig:frequency_sweep_detail}
\end{figure}

As Fig.~\ref{fig:frequency_sweep_overview} shows in the bottom panel both
the baseline and the filter-based algorithm ($\Tcut=25$) can
correctly detect across the entire frequency range, albeit the
baseline lacking robustness at lower frequencies due to its
susceptibility to noise events, and the filter failing at the
transition from 67 kHz to 125 kHz due to the rapid change in the ratio
of ON to OFF events. For details see
Fig.~\ref{fig:frequency_sweep_detail}.

\subsection{Scaling from Single Pixel to Full Sensor}
\label{sec:single_pixel_to_sensor}

As has been shown in section \ref{sec:single_pixel_experiments}, with
aggressive bias settings the SilkyEVCam can be used to detect
frequencies over a range of almost five orders of magnitude. However
this feat is only possible if the ROI is set to a single
pixel. When using the full sensor the dark noise alone already
generates sufficiently many events to exceed the 50 Mevs bandwidth of the sensor
readout, leaving no choice but to revert to more conservative bias
settings. In this section we show that even with default bias settings
sensor readout speed is of critical importance for frequency
detection.

Unfortunately the manufacturer of the sensor (Prophesee) could not
disclose more details regarding internals of the readout electronics
but we observe that when bandwidth saturates, the events delivered by
the SDK are largely in order of monotonically increasing timestamps and with
row-major pixel layout. In other words the events arrive as a dense stream
from the top left to the bottom right image corner. If every pixel fires,
the sensor resolution of 640 x 480 and the bandwidth limit of 50Mev/s
imply that every pixel is updated a little fewer than 163 times per second. By the
Nyquist theorem the maximum detectable frequency is thus about 81hz.

\begin{figure}[t]
  \centering
  \includegraphics[width=1.0\linewidth]{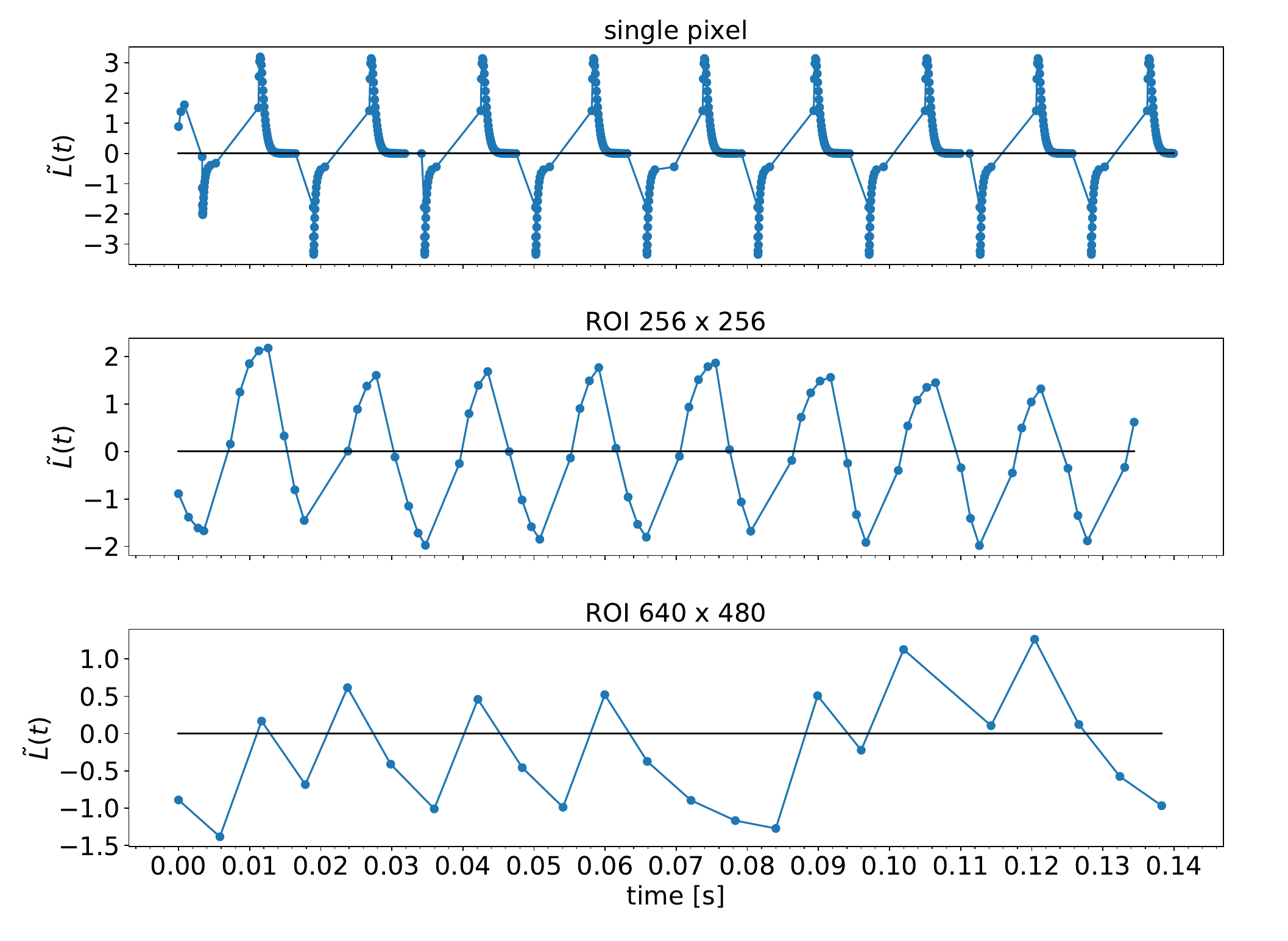}
  \caption{Approximate reconstruction of the brightness of the center
    pixel with varying region of interest (ROI) settings. The signal
    is a square wave at 64Hz (period of 15.625ms). As the ROI
    is scaled from a single pixel (top  panel) to a 256x256 pixel
    patch (middle panel) to full sensor (bottom panel), the signal is
    degraded as increasingly more events are dropped due to readout bandwidth limitations.}
   \label{fig:roi_reconstruction}
\end{figure}

\begin{figure}[t]
  \centering
  \includegraphics[width=1.0\linewidth]{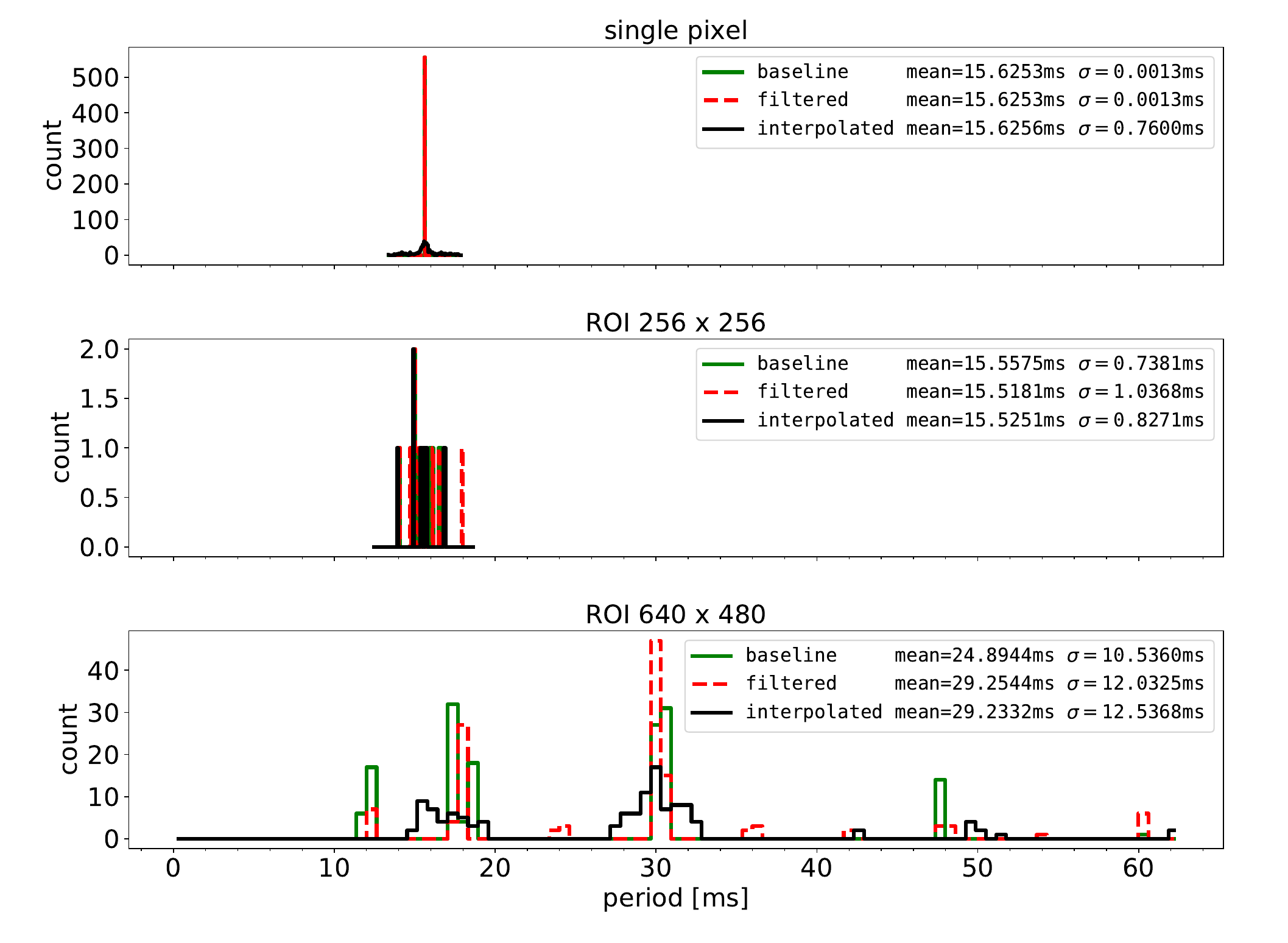}
  \caption{This figure shows how frequency detection deteriorates as
    the ROI is increased from a single pixel (top
    panel) to a 256x256 pixel patch (middle panel) to full sensor
    (bottom panel).  The sensor is illuminated uniformly with a square
    wave of a period of 15.625ms (64Hz) and the filter cutoff is set
    to $\Tcut=25$. The filtering approach recovers the
    mean accurately when the ROI only comprises a single pixel but
    deteriorates as more pixels are readout from the sensor.
    For the full sensor case the frequency detection fails completely
    for baseline, filter, and interpolated filter (c.f. bottom panel
    of Fig.~\ref{fig:roi_reconstruction}).
    Note that for each case the statistics are entirely derived
    from the single pixel at the center.}
   \label{fig:roi_freq}
\end{figure}

In practice the dropping of events due to bandwidth
saturation affect the detection already at lower frequencies since
events do not arrive equidistant in time, in particular for a square
wave signal.  Fig.~\ref{fig:roi_reconstruction} shows an approximate
reconstruction ($\Tcut=25$) of a 64Hz square wave for the center pixel when just
the center pixel is active, when a ROI of 256x256 is configured, or
when the full sensor is active. The bottom panel of
Fig.~\ref{fig:roi_reconstruction} demonstrates why even the most robust of
algorithms considered here (the baseline) fails to detect the correct
frequency. As Fig.~\ref{fig:roi_freq} shows, frequency detection is
already affected at an ROI of 256x256 which comprises just 20\% of the
sensor surface. Note that all statistics shown in
Fig.~\ref{fig:roi_freq} are derived solely from the center pixel,
i.e. the deterioration is due to ``cross talk'' from other pixels as
they compete for readout bandwidth.

\begin{figure}[t]
  \centering
  \includegraphics[width=0.8\linewidth]{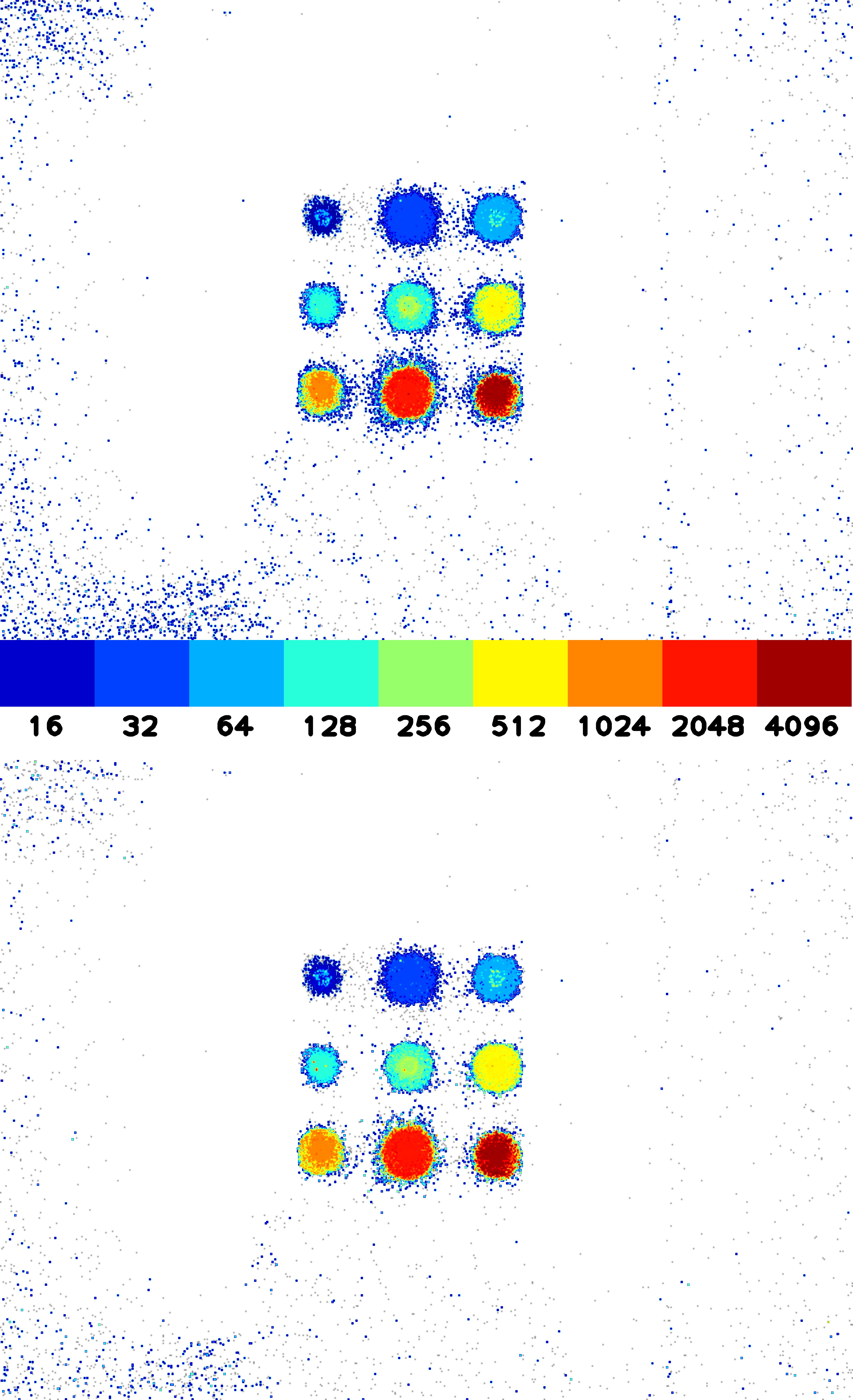}
  \caption{Frequency image of nine LED lights driven with square
    wave signals ranging from 16Hz to 4.1kHz, demonstrating a
    difficult to obtain frequency image. The top image was generated with
    a filter of $\Tcut=5$ whereas the bottom is generated using
    Prophesee's proprietary vibration analysis module. Grey color indicates pixels for which no
    frequency could be determined, but that have had events during the
    readout period (10ms). Note that the visible blur is due to lens
    flare, not lack of focus.}
  \label{fig:leds_freq}
\end{figure}

The sensor readout bandwidth limits thus profoundly affect the
frequency range that can be detected when operating in full sensor
mode. Fig.~\ref{fig:leds_freq} shows an example of the maximum achievable
frequency range when observing LEDs driven by a square
wave. The frequencies start at 16Hz and double every time to reach
4096Hz, thus covering about two orders of magnitude
compared to the almost five orders of magnitude possible in a
single-pixel scenario. For these tests the sensor bias tunings were set to
``fast (LEDs)'' as shown in
Tab.~\ref{tab:bias_settings}. Although only roughly 4\% of pixels are
active and about 2.5\% produce valid frequency detections the readout
bandwidth utilization is already at close to 50\%.

Fairly strong intensity light is required to elicit events at high
frequencies, leading to lens flare that could not be eliminated
despite experimenting with several high quality lenses. The frequency
image from Fig.~\ref{fig:leds_freq} was ultimately obtained with a
Computar M0814-MP2 8mm 1:1.4 lens with fully open aperture.  To reduce
interference between the light coming from different LEDs the
intensity of each LED was adjusted to account for the low-pass
filtering characteristics of the front-end photo receptor
circuits. Note that the image is in focus and the LEDs are all the same
diameter. The size apparent size difference and visible blur are due
to the remaining lens flare mentioned earlier.
We also find lens flare
originating from high-frequency LEDs to interfer with frequency
detection for the low-frequency LEDs as any residual high frequency
signal will be detected while the low-frequency LEDs are switched
off. Thus the high dynamic range of the event camera not only allows it to detect
flickering pixels at widely varying brightness levels but also renders
lens flare much more pronounced in dark regions near a flickering
light.

The top of Fig.~\ref{fig:leds_freq} is obtained by using the filtering
equation \ref{eq:filter_recursion_ab} but no dark noise filtering is
necessary for the ``fast (LEDs)'' bias setting. For comparison
the bottom image shows a frequency analysis using Prophesee's
Metavision toolkit. More details regarding frequency images follow in
Sec.~\ref{sec:visualizing_frequencies}.

\subsection{Visualizing frequencies}
\label{sec:visualizing_frequencies}
Much like a thermal imaging camera fascinates by instantly visualizing
hot and cold objects in the environment, so the event camera can
visualize frequencies over the whole field of view. This is
accomplished by running the  period estimation algorithm from
Sec.~\ref{sec:digital_filter} for each pixel followed by a
frequency image readout step at fixed time intervals. Except for the
necessarily frame-based readout for display, this approach remains fully asynchronous.

As mentioned in Sec.~\ref{sec:single_pixel_to_sensor} using the
whole sensor requires bias settings that severely degrade the range of
detectable frequencies. Under default settings the sensor's front-end
circuit filters out high frequency components of the original signal,
making it less likely to encounter signals of the form shown at the
top of Fig.~\ref{fig:simple_detrend}. Rather, when visualizing
vibrations or periodic motion as they occur in natural scenes most
signals only have a single ON/OFF transition per period and as such
do {\em not} benefit from the use of the digital filter 
described in Sec.~\ref{sec:digital_filter} over the
baseline method. In the context of frequency visualization,
improved frequency accuracy as it can be obtained by
interpolating the time when the signal crosses through zero (cf
Sec.~\ref{sec:digital_filter}) is also irrelevant because such
an improvement is hardly visible in a color-coded frequency image. For
frequency visualization, other criteria matter more:
\begin{itemize}
\item fast detection of a frequency in the acceptable range,
  preferably before even a full period has passed,
  \item rapid timeout detection when a pixel no longer produces
    events, and
  \item holdover of pixels that have missed events for one or more periods.
\end{itemize}
The first two criteria are important to preserve edges of fast moving
objects and to avoid trailing ghost pixels.
To this end we modify and augment the filter-based approach from
Sec.~\ref{sec:digital_filter} as follows:
\begin{itemize}
\item If a pixel has no period assigned to it yet, the time of a zero crossing from
  below or above is used to establish a half period by subtracting the
  time of the previous transition of opposite sign. This will
  establish a period estimate as soon as possible.
\item A period estimate based on a half period will be replaced by
  one from a full period as soon as same-sign transitions are
  available. This means that periods can also be measured between zero
  level crossings from below as opposed to the more accurate
  measurements via zero crossings from above.
\item A pixel's period estimate is timed out (deemed invalid) after no
  events have been received for $n_{\mathrm{timeout}}$ full
  periods. We use $n_{\mathrm{timeout}} = 2$ for all experiments
  presented here.
\end{itemize}
Note that a pixel only times out when period
estimates are read out (the frequency image is generated),
so no extra timers have to be maintained to
remove inactive pixels. Although strictly the baseline method is
sufficient for frequency imaging, we nevertheless deploy the digital
filter with $\Tcut=5$ to show that filter is computationally efficient
and its response is sufficiently fast.

Visualizing periodic signals across the full sensor has attracted
enough industry interest for Prophesee to develop a
closed-source software solution for this. Their MetaVision toolkit
provides the analytics module 
(\verb|FrequencyMapAsyncAlgorithm|) which we will use as a reference for
qualitative comparison. The detailed workings of this module are not
documented, but it only takes two non-obvious parameters:
\verb|filter_length|, which is the number of times the same period must
be measured before outputting an event, and
\verb|diff_thresh_us| which
is the maximum period difference allowed between two consecutive periods to be
considered identical. For the experiments conducted here we set
\verb|diff_thresh_us| to 100us and \verb|filter_length| to one to avoid
trailing ghost images for moving objects.

Fig.~\ref{fig:quad_freq} shows the frequency image of a flying quad
rotor with 6in propellers spinning at about 7500 rpm. Since a
propeller has two blades this translates to frequencies of around
250Hz. The top part of Fig~\ref{fig:quad_freq} is
obtained by our method with $\Tcut=5$, the bottom with Prophesee's
vibration analysis tool at a readout frequency of 100Hz.
The camera is observing the quad rotor from below, with bias settings
as listed in Tab.~\ref{tab:bias_settings} under ``fast (quad rotor)''. Although the analysis
covers the full sensor, for display purposes the image is cropped to
242 x 281 pixels. Evidently our method produces images very similar
to the ones from Prophesee's module. One can clearly see how the quad rotor
is spinning the props at different speeds to generate torque for
attitude control.

\begin{figure}[t]
%python3 ./mv_vs_frequency_cam.py -b ../../event_fourier/bags/guitar/nylon_default_60_slice --freq_min 70 --freq_max 300 -l 73.4 110.0 146.8 185.0 220.0 293.7 --font_scale 10.0 --font_thickness 20 --scale 5.0 --text_height 50
% then pick frame 577
  \centering
  \includegraphics[width=0.8\linewidth]{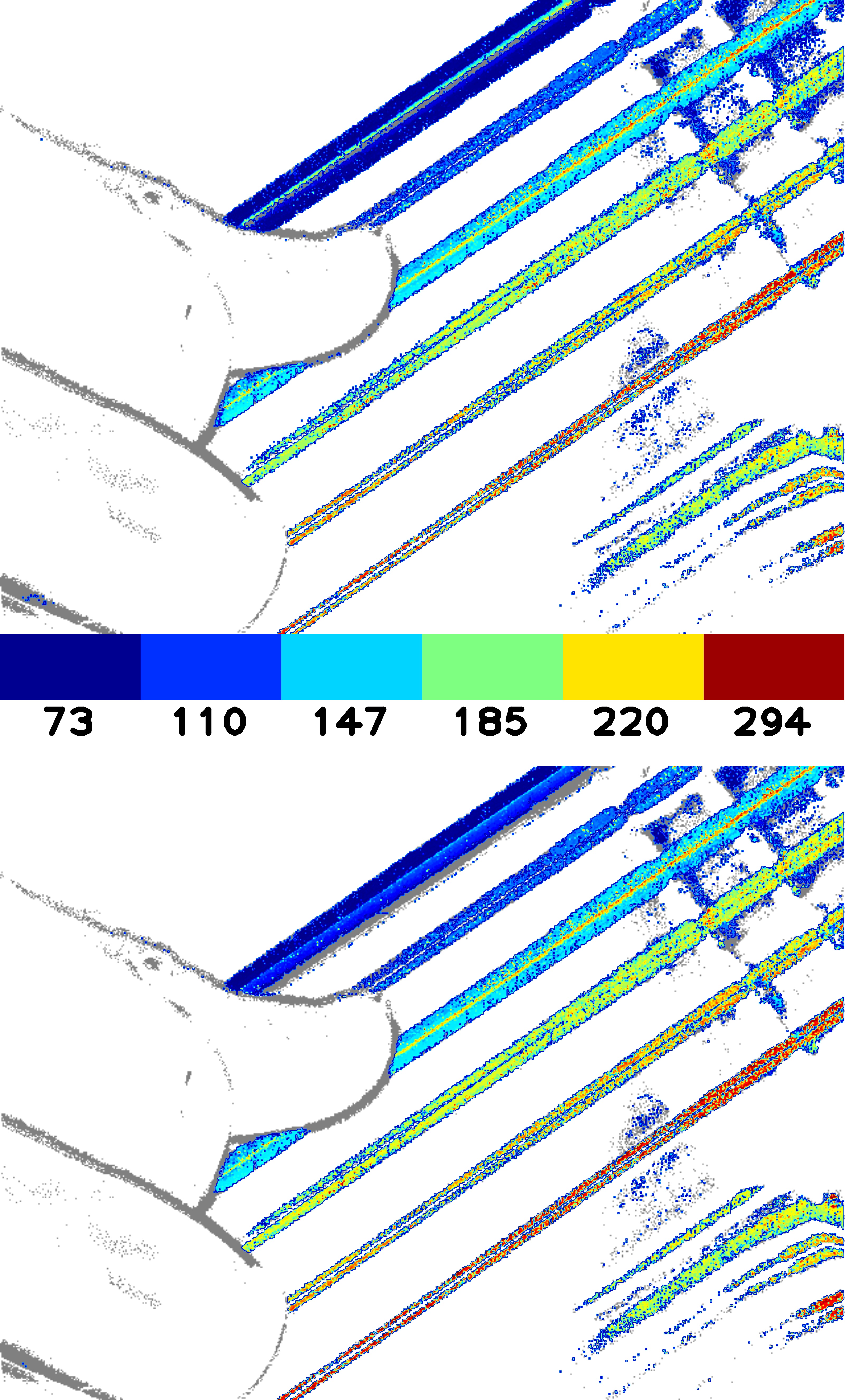}
  \caption{Frequency image of a guitar being played with
    color coded frequencies (in Hz) corresponding to the fundamental modes
    of the strings of an open-D tuned guitar. The top image was generated with
    a filter of $\Tcut=5$ whereas the bottom is generated using
    Prophesee's proprietary vibration analysis module. Grey color indicates pixels for which no
    frequency could be determined, but that have had events during the
    readout period (10ms). The guitar is in open D tuning
    corresponding to the color sample patches. Note the
    frequency doubling at the center of the top most string (low D)
    and the signal from the shadow in the lower right corner.}
   \label{fig:guitar_freq}
\end{figure}

A second demonstration of the algorithm is presented in
Fig.~\ref{fig:guitar_freq}. It shows the vibrating strings of a
nylon string guitar in open-D tuning (D-A-D-F\#-A-D) recorded with the
camera in default bias settings. Again our
frequency image is very similar to the one produced by the Metavision
SDK module albeit we capture the 73Hz frequency of the top most
string more accurately. The thin band of double frequency at the center of
the string is due to the string passing twice through the center
during each period.

While Fig.~\ref{fig:guitar_freq} shows that it is possible to obtain
frequency images of instruments it must be mentioned that detecting
the vibrations of the high note strings requires favorable
experimental conditions. For instance the thin strings of a steel string
guitar produce hardly any signal. Even the nylon strings demand strong
illumination, in this case direct full sunlight at shallow angle with a matte black paper
covering the guitar's sound board to minimize reflection and
shadow. The shadow of the strings is still visible at the bottom right corner of the
image and on the fretboard at the top right.

\section{Runtime Performance and Implementation}
\label{sec:performance}
The digital filter and zero-level crossing detection presented in
Sec.~\ref{sec:digital_filter} can be implemented
efficiently in C++ to run at 75 Mev/s on a laptop class AMD Ryzen 4800H
CPU clocked at 2.9GHz. Since the camera's maximum event rate is 50
Mev/s this allows for real-time frequency imaging under full load. The
filter itself requires only one multiply and two fused 
multiply-add instructions per event which for modern CPUs is a very
modest compute load. When eliminating all memory access by
implementing the filter in registers we find the floating point
arithmetic itself to only take about 2.5ns, i.e an implied rate of
about 400 Mev/s.

However, as is often the case when implementing
neuromorphic algorithms on general purpose CPUs, the real bottleneck
is memory access. The full state required is 21 bytes: two 4-byte float
variables for lagged $\ltil$, one byte for the lagged polarity, two
4-byte float variables for the previous zero level crossing times from
above and below, and one 4-byte float variable with the current period
estimate. Padding inflates the 21 bytes to 24 bytes of state per
pixel, or 7MB total which fits just into the 8MB L3 cache of the CPU. 
Updating the filter state requires two 4-byte load operations for the
lagged $\ltil$ and one 4-byte load for the lagged
polarity and the same number of store operations. When zero-level
crossings are detected additional load/store operations are required to read and update the
times of last zero level crossings and the current period
estimate. Typically these variables are already in cache after the
filter update because of the CPU's 64 byte cache line. Nevertheless
they consume additional cache memory and due to the 
fairly random access pattern of event 
updates performance suffers substantially once the entire 
state no longer fits into the CPU cache. For our particular CPU
however the cache size is sufficient so long as 4-byte float variables
are used as opposed to 8-byte double variables.

Note that memory access would not be
as important an issue for a GPU based implementation due to the typically large
amount of fast memory available there. 

\section{Conclusion}
In this work we explore the task of using an event based
camera to find and visualize the fundamental frequency of
time-periodic signals in a scene.
We show that a combination of
aggressive bias tuning, dark noise filtering, and an efficient
digital IIR filter can reliably detect frequencies across five orders
of magnitude up to 64kHz when restricting the region of interest (ROI) to a single
pixel. We compare our digital IIR filter to a simple baseline method
and determine under which conditions the use of a filter can achieve
more robust and accurate frequency estimates. For full sensor frequency imaging
we find the limited readout bandwidth responsible for greatly reducing
the capabilities of the camera, suggesting substantial benefits for
circumventing bandwidth limitations via an on-camera hardware based
solution. Furthermore in high-contrast scenes we find frequency images
to be marred by strong lens flare due to the high dynamic range of the
camera.

Our open source FrequencyCam imaging algorithm is found to produce
frequency images that are very similar to Prophesee's vibration
analysis module. The full source code for our ready-to-use ROS node
and all data sets used for this paper are accessible under a
permissive license at \url{https://github.com/berndpfrommer/frequency_cam}. 

%
% Acknowledgements
%
\section{Acknowledgements}
The author acknowledges useful and encouraging discussions with
Kenneth Chaney, Ziyun Wang, Tobi Delbruck and Kostas Daniilidis.

%%%%%%%%% REFERENCES
{\small
\bibliographystyle{ieee_fullname}
\bibliography{frequency_cam}
}

\end{document}